# Sequence To Sequence Learning For Unconstrained Scene Text Recognition

Nile University

School of Communication and Information Technology

By Ahmed Mamdouh

August, 2015

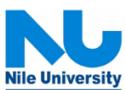

SONY

**Nile University**

**School of Communication and Information Technology**

**Nomination of a Thesis Committee**

Student ID Number:     *0*

Student Full Name:     *Ahmed Mamdouh*

Thesis Title:           *Sequence To Sequence Learning For Unconstrained Scene Text Recognition*

The Program nominates the following persons to serve as the Master's Thesis Committee:

| Name (First, Last) | Center/Program | Academic Rank |
|---|---|---|
| *Dr. Name* /Chair | *School of* | *Professor* |
| *Dr. Name* | *School of* | *Associate Professor* |
| *Dr. Name* | *School of* | *Assistant Professor* |

_______________________________________________     _______________
**Signature of Program Director**                                         **Date**

_______________________________________________     _______________
**Signature of Provost**                                                 **Date**

# Table of Contents

# Acknowledgements

I would like to express my gratitude to, Mark Blaxall and Dr. Fabien Cardinaux, for their supervision and effort. They offered me a great deal of encouragement and their ideas were instrumental to my work.

Thanks also to Dr. Mohamed Elhelw and Dr. Motaz Abdlewahab for giving me the chance to start my research career. I learned a lot from them both.

I am grateful as well to Dr. Motaz Elsaban, the person who first exposed me to the deep learning field during my time in Microsoft ATLC. It was during my time on his team that I learned what research is.

I would like to thank Abubakrelsedik Karali and Ahmed Bassiouny. I remember our discussions fondly and I am blessed to have known them.

Islam Yousry was a tremendous help as well, although that barely does justice to the support he's given me in completing my research and working towards my master's degree. He has been an inspiration since the first day I met him and the presentation he gave about self-driving cars changed my thinking in countless ways.

Thanks to Mohamed Mahmoud Abdelwahb, also known as Fegla, Mostafa Saad, and the whole ACM community, the best I have ever joined. I learned a lot from all of them. They taught me how to write clean and efficient code, how to tackle problems, and even how to think.

Thanks finally to Samasem Elsayed, Mamdouh Abdelkarim, Mohamed, Abdelkarim, and Hager for being such a wonderful family. Words cannot express how grateful I am to all of them for their continuous support.

# Abstract

In this work we present a state-of-the-art approach for unconstrained natural scene text recognition. We propose a cascade approach that incorporates a convolutional neural network (CNN) architecture followed by a long short term memory model (LSTM). The CNN learns visual features for the characters and uses them with a softmax layer to detect sequence of characters. While the CNN gives very good recognition results, it does not model relation between characters, hence gives rise to false positive and false negative cases (confusing characters due to visual similarities like "g" and "9", or confusing background patches with characters; either removing existing characters or adding non-existing ones) To alleviate these problems we leverage recent developments in LSTM architectures to encode contextual information. We show that the LSTM can dramatically reduce such errors and achieve state-of-the-art accuracy in the task of unconstrained natural scene text recognition. Moreover we manually remove all occurrences of the words that exist in the test set from our training set to test whether our approach will generalize to unseen data. We use the ICDAR 13 test set for evaluation and compare the results with the state of the art approaches [11, 18]. We finally present an application of the work in the domain of for traffic monitoring.

## Keywords

CNN: Convolutional Neural Networks
LSTM: Long Short Term Memory
SVM: Support Vector Machines
HOG: Histogram of Oriented Gradient
ICDAR: International Conference on Document Analysis and Recognition
RNN: Recurrent Neural Networks
BPTT: Back Propagation Through Time
FNN: Feedforward Neural Networks
BLSTML: Bidirectional Long Short Term Memory layer
OCR: Optical Character Recognition
SIFT: Scale Invariant Feature Transform
JOINT-CNN: A model that joins the character sequence encoding model with the n-gram model
JOINT-LSTM: A model that joins the output of our proposed model with the n-gram model

# List of Figures

# Chapter 1    Introduction

## 1.1    Motivation

The scene text recognition task is a challenging task, but is very useful and has many practical applications. The scene text recognition task is considerably more challenging than well-structured documents OCR. The difficulty arises due to texture, lighting conditions, and diverse text patterns in terms of font size and types. Moreover complex background could be visually similar to text.

Scene text recognition is widely used in numerous applications. Such a system can help impaired or blind people to read the ingredients of products in the supermarket or menus in restaurants or coffee shops. Also it can be used in cars to read road signs and alert drivers if they break these signs or even can help self-driving cars to follow road signs.  Travelers can use it as an augmented reality application to discover the sights they are visiting by checking them online. Moreover such systems ease human-machine communication like data entry, extracting business card information, making electronic images of printed documents searchable e.g. google books, etc...

Recently, the computer vision research field has seen a shift from using hand crafted features to learned features using Deep Neural Networks (DNNs). DNNs are employed to solve the scene text recognition task. State of the art approaches use convolutional neural network based approaches [11, 12]. Previously, researchers were using hand crafted features with sophisticated models such as conditional random fields (CRFs) [29] and pictorial structure models [30].

## 1.2    Objectives

In this thesis, we describe a novel approach that employs techniques from natural language processing to tackle the unconstrained scene text recognition problem. Since the CNN-based approach lacks the use of contextual information, the sequence to sequence learning using a Long Short-Term Memory network aligns very well with the unstructured scene text recognition task. Even approaches that detect n-gram and character sequence using CNN based models and combine them to make use of contextual information do not solve the visual similarities between characters. Both character sequence and n-gram CNN-based models start from the image and both of them may confuse "0" with "O" or "g" with "9". We found that the best way to solve this problem is to consider the whole word or to take most of the characters into account.

Our model uses a character sequence encoding model and corrects the mistakes using a Long Short Term Memory model. Since LSTMs are good in learning sequences as shown in the sequence to sequence learning with neural networks [26], we have decided to use the LSTM to learn the mapping between the character probability maps generated by the CNN-based model "character sequence encoding" to a correct English word. For example if the CNN detects "sony" as "s0ny", we teach the LSTMs that "s0ny" should be detected as "sony".  Our aim is to teach the LSTM to map the false positives to empty characters, add the missing ones, and replace the incorrect characters with the correct ones.

Yet another of the objectives is to present another way for handling the arbitrary length to length sequence learning with LSTMs.

Our approach achieved state of the art performance on the ICDAR 2013 benchmark and generalizes over unseen words in the training set, as will be shown in the experiments chapter.

This thesis is organized as follows. We start with a literature survey and description of relevant work in Chapter 2. In Chapter 3 we review state-of-the-art approaches using CNNs and LSTMs. Chapter 4 presents our model in detail. In Chapter 5 we present our experiments and show how the LSTM corrects the CNN-based approach output and that LSTMs can generalize and predict words which do not exist in the training set. In Chapter 6 we present traffic monitoring as one of the scene text recognition application. Finally Chapter 7 summarizes the key results and provides ideas for future work.

# Chapter 2    Background

The problem of text recognition from scanned documents is extensively researched and there are many well performing available systems. Text recognition from natural images is less developed and more challenging. The difficulty arises due to the text regions in natural images having no structure and possibly a complex background. Additionally there is wide variability in text fonts, styles, colors, scales, and orientations. Numerous environmental effects such as contrast, shadow, lighting and occlusion of objects from the background make it even more challenging.

Many approaches have been proposed for scene text recognition, including mid-level representation, multi scales mid-level representation [9, 10], global representation using low level hand crafted features such as HOG[5] or SIFT[6], aggregated with bags of words[7] or fisher vector[8] and apply SVM on top of these global representations[1, 2, 3, 4].

Some approaches use lexicon [31, 32] for the word recognition task. Either they limit the lexicon to a specific set of words or include all English words but in both cases they don't handle common names, places, abbreviations, and so forth. Such restriction makes the problem much easier, because if the model makes a mistake in one or two characters the correct word still can be correctly detected by searching the detected word in the list of the given words. The bigger the set of words in the lexicon, the harder the task is. Some applications require lexicon based approach due to the possible words are limited such as food names or street names in a specific town or reading menus in a restaurant.

Traditionally, people calculate hand crafted features for regions of interest and feed them to a classifier to score them for instance, histogram of oriented gradient. Histogram of Oriented Gradient is an engineered feature representation first introduced by Dalal and Triggs [5]. The algorithm describes low level image features while being invariant to geometric transformation within the object's structure. The algorithm divides the image into small connected cells and for each cell counts occurrences of gradient orientation.

One current trend in machine learning is that the feature representation is learned using deep convolutional neural networks. We implement a convolutional neural network and long short term memory-based approach for unconstrained scene text recognition. The CNN model is for feature design and character detection using a softmax layer and the LSTM is for correcting the CNN mistakes such as missing characters, detecting non existing ones and confusing a character with some other characters due to the visual similarities. The convolutional neural network's architecture consists of four convolutional layers, two fully connected dense layers followed by softmax for classification and an input layer. Image patches of a fixed size are the input to the first layer through the input layer, the output of each convolutional layer is a feature map and the output of the $i^{th}$ layer is the input to the $i^{th} + 1$ layer. The last convolutional layer is connected to a logistic regression for classification like support vector machines [17], or softmax. Normally we can intertwine a subsampling, max-pooling, or normalization layer between two convolutional layers. During the training phase all parameters are optimized using a stochastic gradient descent to minimize the classification loss over the training set.

## 2.1 Convolutional Neural Networks (CNNs)

Convolutional neural networks are used to learn deep features from the raw images and followed by softmax classifier to make decisions for the class of each patch. In the scene text recognition task the class $c$ may belong to a…z, 1…9 or Non-Text class, n-gram, or word index in the all English words.

## 2.2 CNNs Features

We use the convolutional neural network to learn deep features which had been proven to perform much better than the hand crafted ones in numerous computer vision problems for example in the object detection problem, the deep features-based approach: rich hierarchy convolutional neural networks [20] by far outperforms the best performing histogram of oriented gradient-based approaches [21].

Let us first introduce what is feature representation and why do we need another representation for images, patches, or objects, rather than just using the RGB values of 3-channel images, gray-level intensity values of gray images or binary values of binary images. The geometrical interpretation of feature vector is a point in N-dimensional space where N is the feature vector size and number of features. Image of size $30 * 30 * 3$ is one point in 2700-dimensional space, each pixel is one dimension. Consider two image patches $p_1$ and $p_2$ of the same number of rows $R$, and number of columns $C$. Patch $p_1$ starts in image $I$ at position $(0,0)$ and ends at $(R, C)$, and patch $p_2$ from the same image $I$ but this time we shift $p_2$ just one pixel to the right so it starts at position $(0,1)$ and ends at image $(R, C + 1)$.

In real life computer vision applications, the distance between $p_1$ and $p_2$ should be very small, but this is not the case here because the first dimension of $p_1$ will be compared with the first dimension of $p_2$ which are not the same, so the distance between two image regions from the same image $p_1$ and $p_2$ may be too large due to very small shift. Also, small changes in the illumination of the same image or image regions will have a drastic effect at the pixel level. The raw pixel feature representation is also sensitive to image scale.

The feature extraction is one of the most important tasks in solving computer vision problems like object detection, scene classification and object tracking. How to extract descriptive features for objects is still an open research area. The feature representation for objects should be invariant to scale, rotation, translation and illumination conditions. Moreover, feature representations that can handle intra-class variation and background clutter are considered successful feature vectors. Feature representations are usually divided into three categories: low-level, mid-level and high-level (also known as holistic features). Features computed at the pixel level like edges, lines, corners, SIFT; HOG, etc. are low-level features. Mid-level features are to describe objects or images at the patch level. The patches should be discriminative and representative, a) to be discriminative, they should not be common in all objects, b) to be representative, they should be occurring frequently in the same objects. For example, bag of visual words and unsupervised discovery of mid-level features [13]. High-level features are to describe the whole object like convolutional neural network features.

The first layer in a CNN learns low level features, like edges from the raw pixels in the input image. Each successive layer learns more complex and more high-level features from the previous one [19] Convolutional neural networks can be seen as normal neural networks, but each node is a convolutional filter, so it can learn spatial features from the local connectivity. Let us consider images with dimensions $32 * 32 * 3$. In normal neural networks each pixel is connected to all neurons in the next hidden layer so the number of weights for each hidden layer is $32 * 32 * 3$. However, for convolutional neural networks each chunk of, for example, $5 * 5 * 3$ is connected to all neurons in the next hidden layer. Each convolutional layer contains multiple hidden neurons. For a normal neural network with 30 neurons in the first hidden layer, the number of weights is $30 * 32 * 32 * 3 = 92160$. Let us assume we apply the convolutional filters at stride 1 for the convolutional filters neural network. We either use weight sharing where we use the same filter everywhere in the image for each neuron in the hidden layer or we use different filters for each $5 * 5 * 3$ region in the input image. The number of parameters without parameters sharing is $5 * 5 * 3 * 30 * 28 * 28 = 1.7M$, where $5 * 5 * 3$ is the filters' dimensions, 30 is number of neurons in hidden layer, and $28 * 28$ is the output dimension after convolution $\frac{N-F}{S} + 1$ where $N$ is the input dimension length, $F$ is the filter size, and $s$ is the stride, so $\frac{32-5}{1} + 1 = 28$. The number of parameters with parameter sharing is $5 * 5 * 3 * 30 = 2250$, where $5 * 5 * 3$ are the filter's dimensions and 30 is number of neurons in the hidden layer. With parameter sharing the number of parameters is dramatically reduced.

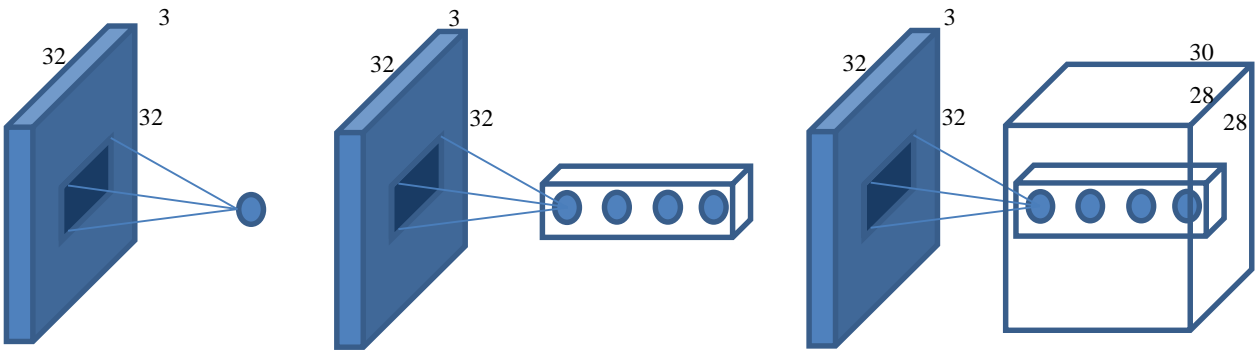

Figure 2:1 How convolutional layers work.

## 2.2.1 Convolutional Layer

Convolutional layers consist of filters that produce feature maps from the image as shown in figure 2:1. Edge detector is an example how these filters look like. The filters can be anything. However edge detector is just an example. The filters weights are learned using the back propagation algorithm. Filters are convolved with the input images and feature maps that have been generated from the previous layer. The convolution for image $I$ by filter $K$ is given by

$$s[i,j] = (I * K)[i,j] = \sum_{m}^{M} \sum_{n}^{N} [i+m, j+n] K[m,n]$$

Where $M$ and $N$ are the filter's dimensions. These filters learn the common and most important features that describe a specific class and discriminate it from the others. The first layer filters learn from the input images low level features like, edges, fine curves, etc... The second layer filters learn more advanced features like shapes T, L, <, >, ^, v. The third learns even more complex features. Actually very complex features like human faces can be learned. The visualizing and understanding convolutional networks paper [19] shows that the deeper we go the more complex feature CNNs can learn.

## 2.2.2 Pooling Layer

Normally we intertwine pooling layer between two convolutional layers as in figure 2:3. Pooling layers help to reduce the number of model parameters by summarizing the output value and its neighborhood outputs within a rectangle window. Pooling layers can replace the output value and its neighborhood output by the maximum for max-pooling or average for average-pooling as shown in figure 2:2.

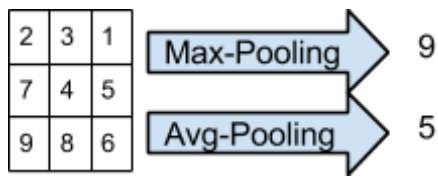

Figure 2:2 pooling layer.

The pooling layers make the generated feature maps tolerate fine displacement changes. In other words, help to generate local displacement invariant feature maps.

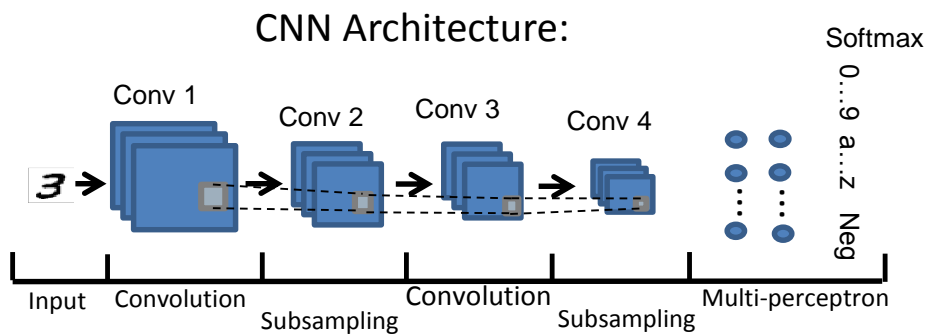

Figure 2:3. Feature maps after each CNN and pooling layer.

## 2.3   Logistic Regression - Softmax

After feature vectors are extracted for the images or image patches, a classifier is trained on a training dataset of different classes. The classifier identifies which class a new image or patch belongs to.

The simplest way is to measure the distance between the new image feature vector and the other training image feature vectors and assign the label of the closest training image. This is a supervised classification approach named K-Nearest Neighbors (K-NN) where $K$ should be chosen by cross-validation. In K-NN, all training feature vectors are compared against the testing samples, which is very expensive in terms of time and memory as the classifier has to remember all training data.

The problems of supervised and unsupervised classification have been intensively studied and many classifiers have been introduced like KNN, SVM, Naive Bayes, soft-max, logistic regression and decision trees. Usually, the number of training data, available memory, and the testing time define which classifier is most suitable. Also if the data is not linearly separable, it's better to choose a non-linear kernel function.

A simple probabilistic classifier like naive Bayes takes linear time to train, but it assumes the features are independent. Logistic regression doesn't assume conditional features independence. So, it can learn something like if someone likes ice cream in winter but not in summer. Logistic regression gives probabilistic interpretation like naive Bayes. Moreover, in logistic regression you can update your model for a new data item without having to repeat the whole training process. On the other hand, logistic regression is slower than naive Bayes and doesn't perform very well in high dimensional feature vectors.

Support vector machines (SVMs) unlike logistic regression, computes uncalibrated and hard to interpret scores for each class, but they perform very well in high dimensional feature vectors.

Soft-max is one of a probabilistic discriminative model that takes an input vector $x$ and assigns it to one of $C$ classes we shall start with binary class classification first and extend it to the general form.

## 2.3.1 Binary Classification

The simplest form for binary classification is

$$y(x) = w^t x + w0$$

Where $w$ and $w_0$ are the model's parameters and the bias respectively. The input vector $x$ belongs to class $C_1$ if $y(x) \geq 0$ and to $C_2$ otherwise. The decision boundary is defined by $(D-1)$-dimensional hyperplane when $y(x) = 0$, where $D$ is the input vector $x$ size. All points lie on the decision boundary $y(x) = w^t x + w_0 = 0$, for both $x_a$ and $x_b$ lie on the decision boundary $y(x_a) = w^t x_a + w_0 = 0$ and $y(x_b) = w^t x_b + w_0 = 0$, so $w^t(x_a - x_b) = 0$ therefore, the vector $w$ is orthogonal to all vectors lying within the decision boundary hyperplane. Since $w_0$ determines the location of the decision boundary we can measure the normal distance from the decision boundary to the origin by $\frac{w^t}{\|w\|} = -\frac{w_n}{\|w\|}$.

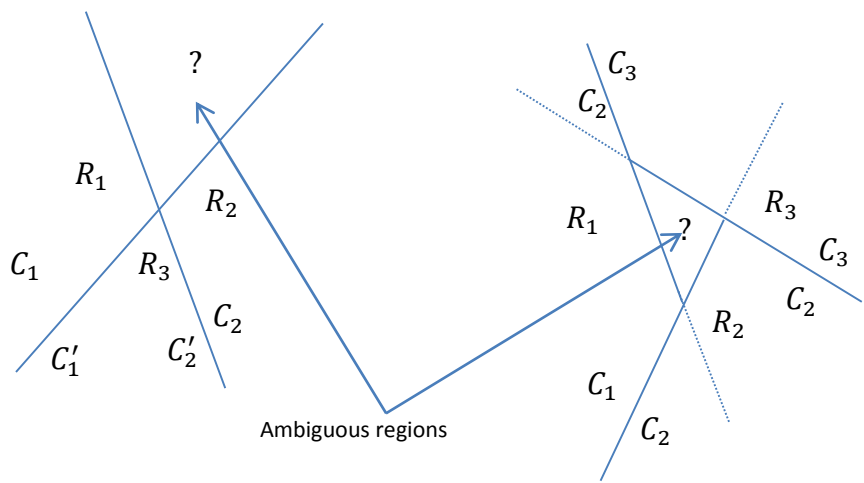

Ambiguous regions

Figure 2:4 shows the ambiguous regions.

## 2.3.2 Multiclass Classification

Now consider the multi-class case where $C_k$ takes $K$ values from 0 to $K-1$ and $K > 2$. We can make use of the binary classification to solve the multi-class version in two ways.

a) We can use $K - 1$ binary classifiers each of which classifies points from a particular class $C_k$ of the other points not in that class. That is known as one-versus-the-rest classification.

b) The other one is to use $\frac{K*(K-1)}{2}$ binary classifiers, one for every possible pair of classes.

Both a) and b) introduce ambiguous regions where we could not figure out to which class a point in this regions belongs to, as shown in figure 2:4. Alternatively, we can avoid the ambiguous regions by considering $K$-class discriminant consists of $K$ linear functions of the form

$$y_k(x) = w_k^t + w_{k0}$$

A point $x$ is assigned to class $C_k$ if $y_k(x) > y_i(x)$ for all $k \neq i$. The decision boundary between $C_k$ and $C_i$ is given by $(D - 1)$-dimensional hyper-plane when $y_k(x) = y_i(x)$.

Soft-max is a model that generalizes logistic regression which can be used in multi-class classification problems. So, it perfectly fits to our text detection problem where we have 37 different classes: 26 characters, 10 digits, and 1 negative class to handle the background patches.

Let's first introduce how logistic regression works on binary classification where $C_k$ takes either 0 or 1. Logistic regression uses iterative approach to approximate the model parameters $w$ of size $M$ directly, where $M$ is the number of features. We can estimate the model parameters indirectly using generative models by fitting class-conditional densities and class priors separately, then apply Bayes' theorem. The number of model parameters if we had fitted Gaussian class conditional densities is $\frac{M(M+5)}{2} + 1$. Since we have two classes we should estimate $2M$ parameters for the means, $\frac{M(M+1)}{2}$ parameters for the shared covariance matrix and the class prior $P(C_1)$. The number of model parameters of logistic regression is linear in M compared to quadratic in the generative models. We shall show what the model parameters of logistic regression are and how to estimate them using maximum likelihood.

The logistic sigmoid function is the activation function of logistic regression $p(C_1|x) = y(x) = \sigma(w^t x)$ where $\sigma(a)$ is the logistic sigmoid function $\sigma(a) = \frac{1}{1+e^{(-a)}}$

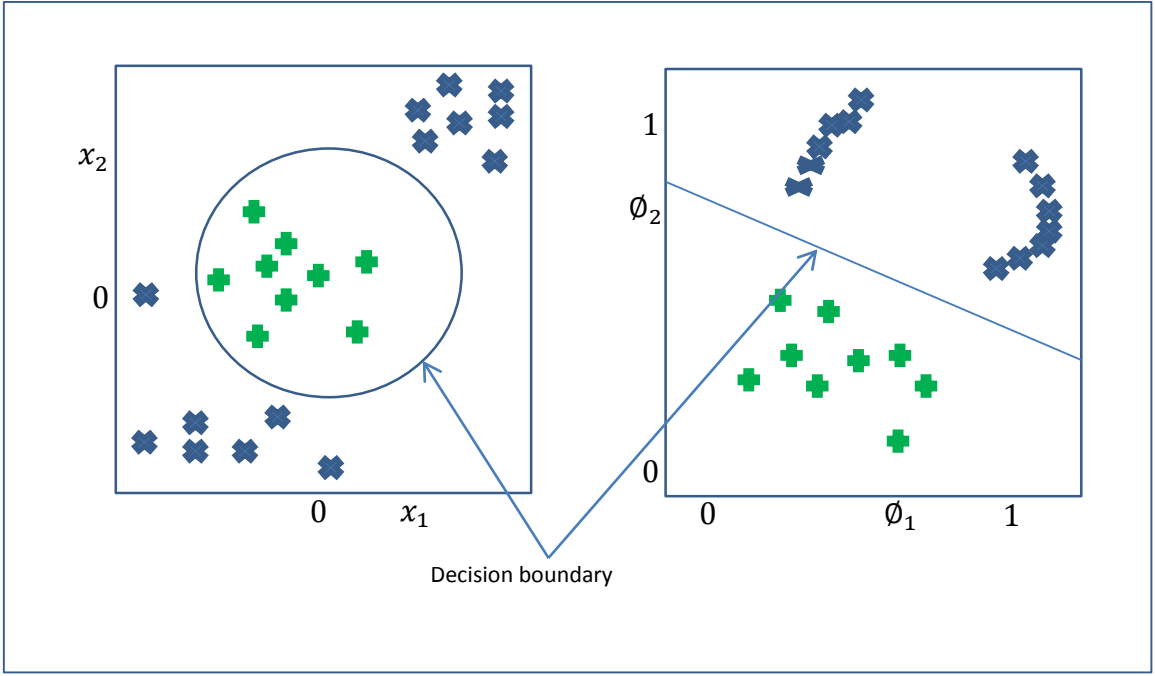

Figure 2:5 illustrate the role of nonlinear basis function.

Now we estimate the logistic model parameters $w$ using maximum likelihood. Consider a dataset $\{\emptyset_n, t_n\}$ where $t_n$ is the class label of the $n^{th}$ input feature vector $x_n$ and $\emptyset_n$ is nonlinear basis function as shown in figure 2:5. The likelihood function is

$$p(t|w) = \prod_{n=1}^{N} y_n^{t_n} \{1 - y_n\}^{1-t_n}$$

Where $t = (t_1, \ldots, t_n)^t$ and $y_n = p(C_1|n)$ . We can maximize the likelihood by minimizing the negative of log-likelihood.

$$E(w) = -\ln p(t|w) = -\sum_{n=1}^{N} \{t_n \ln y_n + (1 - t_n) \ln(1 - y_n)\}$$

Where $y_n = \sigma(a)$ and $a = w^t x_n$. Then we take the derivative of the logistic sigmoid function $a$ with respect to the model parameters $w$.

$$\frac{\partial \sigma}{\partial a} = \sigma(1 - \sigma)$$

$$\nabla E(w) = -\sum_{n=1}^{N} \{t_n \frac{1}{\sigma_n} \sigma_n (1 - \sigma_n) \emptyset_n + (1 - t_n) \frac{1}{1 - \sigma_n} (-\sigma_n (1 - \sigma_n)) \emptyset_n\}$$

$$\nabla E(w) = -\sum_{n=1}^{N} \{t_n (1 - \sigma_n) \emptyset_n + (1 - t_n) \sigma_n \emptyset_n\}$$

$$\nabla E(w) = -\sum_{n=1}^{N} (y_n - t_n) \emptyset_n$$

The difference between predicted output $y_n$ and the target value $t_n$ times the nonlinear basis function $\emptyset_n$ for input feature vector data $x_n$ contributes to the gradient update for model parameters $w$. Iteratively, the optimal $w$ can be found using gradient descent

$$w^{t+1} = w^t - \eta \frac{\partial E}{\partial w^t}$$

Where $\eta$ is the learning rate. The model parameters of the next iteration $w^{t+1}$ depend on the current model parameters $w^t$, learning rate and the partial derivative of the error by the model parameters $\frac{\partial E}{\partial w^t}$.

Now consider the case of multiple classes where $C_k$ takes $K$ values from 0 to $K-1$ and $K \geq 2$. We use the soft-max function to handle the multi-classes problem, softmax is given by

$$p(C_k|\emptyset) = y_k(\emptyset) = \frac{e^{a_k}}{\sum_j^K e^{a_j}}$$

Where $a_k$ is given by $a_k = w_k^t \emptyset$. Then determine the model parameters $w$ using maximum likelihood. The simplest way to extend the maximum likelihood from two classes form $p(t|w) = \prod_{n=1}^N y_n^{t_n} \{1 - y_n\}^{1-t_n}$ to the general form of multiple classes is to encode the target vector $t_n$ using the 1-of-k encoding scheme. Then we can write the general form of the maximum likelihood as follow

$$p(T|w_1,\ldots,w_k) = \prod_{n=1}^N \prod_{k=1}^K y_{nk}^{t_{nk}}$$

Where $T$ is the predicted target values, $K$ target values for each data point $x_n$ and $y_{nk} = y_k(\emptyset)$. Similarly in the two and multi-class cases, we can maximize the likelihood by minimizing the negative of log-likelihood

$$E(w_1,\ldots w_k) = -ln\, p(T|w_1,\ldots w_k) = \sum_{n=1}^N \sum_{k=1}^K t_{nk} ln\, y_{nk}$$

This is known as the cross entropy error function. We now take the derivative of $y_k$ with respect to $a_j$ to get.

$$\frac{\partial y_k}{\partial a_j} = y_k(I_{kj} - y_j)$$

Where $I$ is the identity matrix. We can obtain the gradient update by applying the derivative on soft-max function with respect to the model parameters $w$.

$$\nabla w_j E(w_1,\ldots w_k) = -\sum_{n=1}^N (y_{nj} - t_{nj})\emptyset_n$$

Similar to the binary case, the difference between predicted output $y_{nj}$ and the target value $t_{nj}$ times the nonlinear basis function $\emptyset_n$ for input feature vector data $x_n$ contributes to the gradient update for model parameters $w_k$. Iteratively, the optimal $w_k$ can be found using gradient descent

$$w_k^{t+1} = w_k^t - \eta \frac{\partial E}{\partial w_k^t}$$

Where $\eta$ is the learning rate. The model parameters of the next iteration $w_k^{t+1}$ depend on the current model parameters $w_k^t$, learning rate $\eta$ and the partial derivative of the error by the model parameters $\frac{\partial E}{\partial w_k^t}$.

## 2.4    Long Short-Term Memory - LSTM

LSTM is a recurrent neural network that can remember long sequences without causing the gradient to vanish or explode as shall be explained shortly in detail.

### 2.4.1  Recurrent Neural Networks

Recurrent neural network is a network with neurons and feedback connections that can learn arbitrary sequences. It has internal states $s_t$,

$$s_t = F_\theta(s_{t-1})$$

Where $F_\theta$ is the mapping function that maps the state $s_t$ at time $t$ to the state at $s_{t-1}$ at time $t-1$.

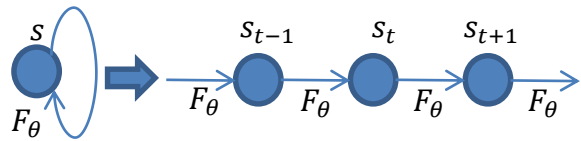

Consider another system with an external input $x_t$

$$s_t = F_\theta(s_{t-1}, x_t)$$

The state $s_t$ at time $t$ implicitly encodes information about the whole sequence in the past $(x_t, x_{t-1}, \ldots, x_2, x_1)$.

$$s_t = G_t(x_t, x_{t-1}, \ldots, x_2, x_1)$$

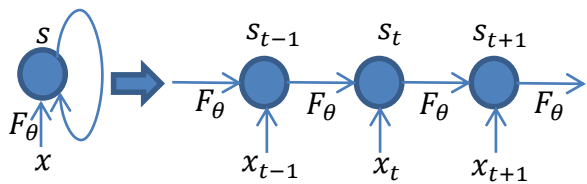

Consider the following recurrent neural network model for a classification problem that maps same length sequence to sequence,

$$a_t = b + W s_t + U x_t$$

$$s_t = \tanh(a_t)$$
$$o_t = c + Vs_t$$
$$p_t = softmax(o_t)$$

Where $b$ and $c$ are the bias vectors, $U, V,$ and $W$ are the input-to-hidden, hidden-to-output, and hidden-to-hidden matrices respectively. The error loss for a sequence is the summation of the losses over all time steps.

$$L(x, y) = \sum_t L_t = \sum_t -\log p_{yt}$$

Where $y_t$ is the target output of input $x_t$ at time step $t$. We can calculate the gradient on parameters $c, b, W, and V$ using Back-Propagation through Time (BPTT) algorithm.

$$\nabla_c L = \sum_t \nabla_{o_t} L \frac{\partial o_t}{\partial c} = \sum_t \nabla_{o_t} L$$

$$\nabla_b L = \sum_t \nabla_{s_t} L \frac{\partial s_t}{\partial b} = \sum_t \nabla_{s_t} L \, diag(1 - s_t^2)$$

$$\nabla_V L = \sum_t \nabla_{o_t} L \frac{\partial o_t}{\partial V} = \sum_t \nabla_{o_t} L s_t'$$

$$\nabla_W L = \sum_t \nabla_{s_t} L \frac{\partial s_t}{\partial W} = \sum_t \nabla_{s_t} L \, diag(1 - s_t^2) s_{t-1}'$$

Where $\nabla_{s_t} L$ refers to the full influence of $s_t$ through all paths from $s_t$ to $L$, $\frac{\partial s_t}{\partial W}$ or $\frac{\partial s_t}{\partial b}$.

The recurrent neural network states may encode information about the whole sequence. The output at time $t$ depends not only on the information captured from $x_1, \ldots, x_t$, but also on some information from the future or the whole input sequence. This type of recurrent neural networks is called bidirectional recurrent neural networks [22]. Bidirectional neural networks were first invented by Schuster and Paliwal, 1997 and have been used for applications such as hand writing [23] (Graves eta al., 2008; Graves and Schmidhuber, 2009), speech recognition [24] (Graves and Schmidhuber, 2005; graves et al., 2013) and bioinformatics [25] (Baldi et al., 1999). The recurrent neural networks can either learn sequences of fixed size like Hopfield networks (Hopfield, 1982) or Boltzmann machines (Hinton, Sejnowski, & Ackley 1984), or time-varying sequences. Recurrent neural networks can suffer from exponential decay of gradient information due to a limitation in the learning approaches. The gradient based approaches "Back-propagation Through Time" (Williams & Zipper, 1992; Werbos, 1998) and "Real-Time Recurrent Learning" gradient evolution exponentially depends on the magnitude of the weights (Hochreiter, 1991) which means the back-propagated error either vanishes or explodes (Hochreiter & Schmidhuber, 1997; Bengio, Simard, & Fransconi, 1994).

## 2.4.2 Constant Error Carrousels

The long-short term memory networks are recurrent neural network that solve the long term dependencies problems. The problem with the recurrent neural networks is that the gradients propagated over many stages tend to either vanish or explode. The LSTM algorithm overcomes these problems by enforcing non-decaying error flow which allows it to bridge minimal time lags in excess of 1000 discrete time steps (Hochreiter & Schmidhuber, 1997) using what so called constant error carrousels (CECs).

The basic unit of the LSTMs is the memory block. It replaces the state $s_t$ unit in recurrent neural networks. The memory block contains one or more memory cells, additive and or multiplicative (forget gate) gates as shown in figure 2:6. Each memory cell has a self-connected linear unit called "Constant Error Carousel" (CEC) which solves the gradient vanishing and explosion problems. The loss of the $j^{th}$ memory block $\delta_{out_j}$ is the summation over all cells $v$ of size $s_j$ in block $j$.

$$\delta_{out_j}(t) = f'_{out_j}(net_{out_j}(t))\left(\sum_{v=1}^{s_j} h\left(s_{c_j^v}(t)\right)\sum_k w_{kc_j^v}\,\delta_k(t)\right)$$

The weights of the input and forget gate can be updated as follow

$$\nabla w_{c_j^v m}(t) = \alpha e_{s_{c_j^v}}(t)\frac{\partial s_{c_j^v}(t)}{\partial w_{c_j^v m}}$$

Where $e$ is the internal memory error.

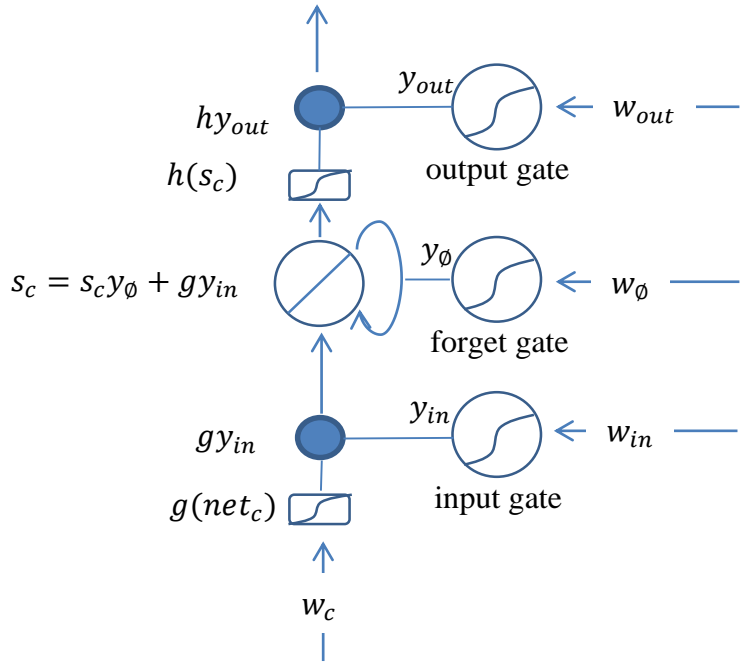

Figure 2:6  Memory block with one cell and a multiplicative forget gate.

# Chapter 3    State-of-the-Art Approaches

The best performing algorithms today for many vision tasks, and text recognition in particular, are based on deep Convolutional Neural Networks. Recent approaches use CNNs to generate probability maps for characters and n-grams [11]. Yet another deep convolutional neural network approach for detecting unconstrained text [12] combines two convolutional neural networks-based approaches. The first one is for character sequence detection and the second for n-grams detection. Both CNN-models start from the images and this is computationally expensive. Approach introduced in this thesis uses one CNN-model, the character sequence encoding, and one LSTM-model which make our approach faster than [12]. Both approaches [11, 12] are much related to our work. We use a convolutional neural network model to generate character sequence maps. Then we use Long-Short Term Memory to correct the output of the convolutional neural network. In this work we use a state of the art approach Synthetic data and artificial neural networks for natural scene text recognition [18] and show how LSTMs improve the performance of the CNN-based model and even beat the state of the art [12] performance.

The following three models are presented in [18] for scene text recognition. The first model in figure 3:1 uses a lexicon of large size where they employ almost all English words; 90K in total. The second is for character sequence encoding. The model produces probability maps for all classes for each character. The third model is trained to spot n-grams using the fully connected layer of size 10,000 nodes followed by softmax.

The current state of the art approach employs the two convolutional neural network models, the character sequence and the n-grams encoding models [12] using conditional random field.

## 3.1    Lexicon-Based CNN Model

The model encodes words from a predefined dictionary. The dictionary covers around 90K English words, including different forms for the same word.  Number of classes equals to number of word, which might seem too large for a classification problem, but the authors perform incremental training to handle such huge number of classes. The model architecture as shown in figure 3:1 consists of four convolutional layers and two fully connected layers. Rectified linear units are used throughout after each weight layer except the last one. Max-pooling layer of size $2x2$ follows the first, second and third convolutional layers. The convolutional layer filters' dimensions are $5x5x64$, $5x5x128$, $3x3x256$ and $3x3x512$ for the first, second, third and fourth convolutional layers respectively.

The evaluation of the lexicon based model for ICDAR 2003 and SVT datasets obtained by finding the lexicon word with the minimum edit distance to the predicted character sequence.

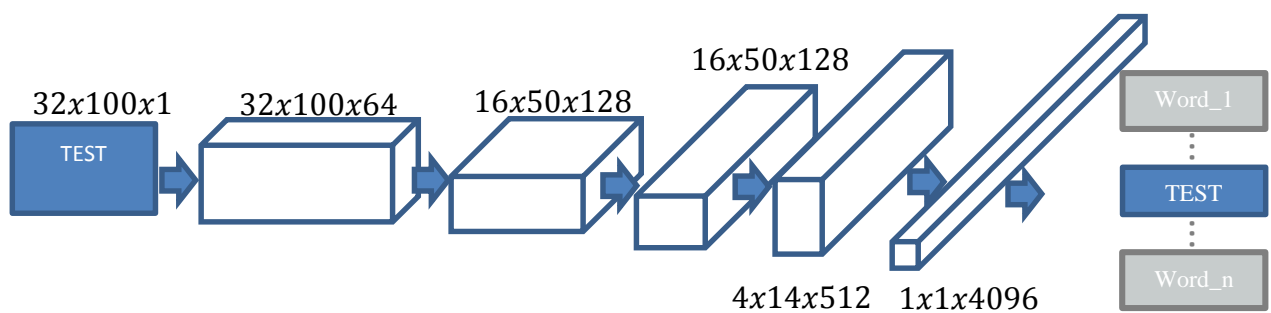

Figure 3:1 Shows the CNN architecture for a dictionary based model.

## 3.2    Character Sequence Encoding

The character sequence encoding model does not have the lexicon restriction and detects sequence of characters by generating the probabilities of the 37 classes (0-1, a-z, and the negative class) for each patch as shown in figure 3:2. The same architecture as described in section 3.1 is used for the character sequence encoding except the last layer. Instead they feed the output of the fully connected layer to 23 fully connected layers of size 37 neurons for each, where the 37 neurons represent number of classes and 23 is maximum word length.

The character sequence model is trained on image samples from the 90K words dictionary. The character sequence is counted as a hit if it perfectly matches with ground truth, otherwise it will be counted as miss.

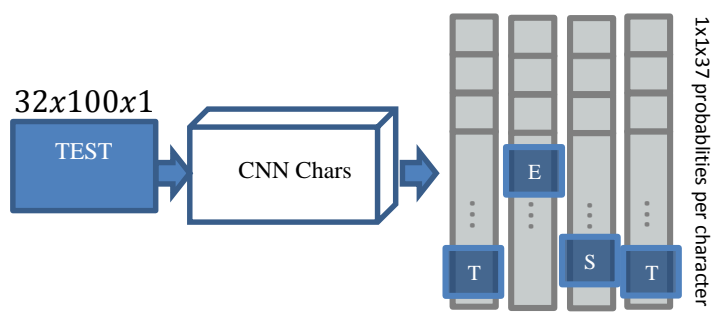

Figure 3:2 shows CNN architecture for a character sequence encoding model.

## 3.3    N-gram Encoding

The third model (shown in figure 3:3) generates all n-grams using the same CNN architecture in figure 3:1, but the final fully connected layer has 10K neurons to map all n-grams presented in the input image after applying the logistic regression function.  Since the word can be presented as a composition of an unordered set of characters n-gram, the CNN is learning to recognize the presence of each n-gram within the input image.

Due to the frequency of n-grams is varying (some n-grams occur more frequent than others) the gradient for each n-gram is scaled by the inverse frequency of their appearance in the training word corpus.

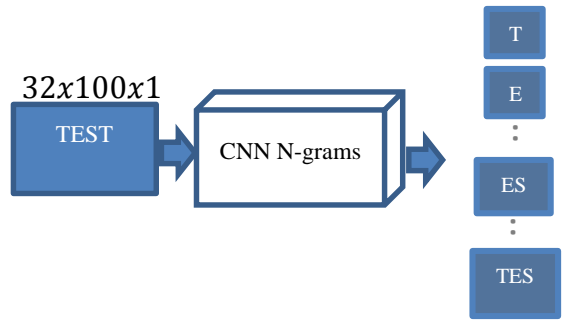

Figure 3:3 shows CNN architecture for the n-gram encoding model

## 3.4 The Joint Model

The state of the art for the unconstrained scene text recognition task joins the character sequence and n-gram encoding models using a conditional random field as shown in figure 3:4.

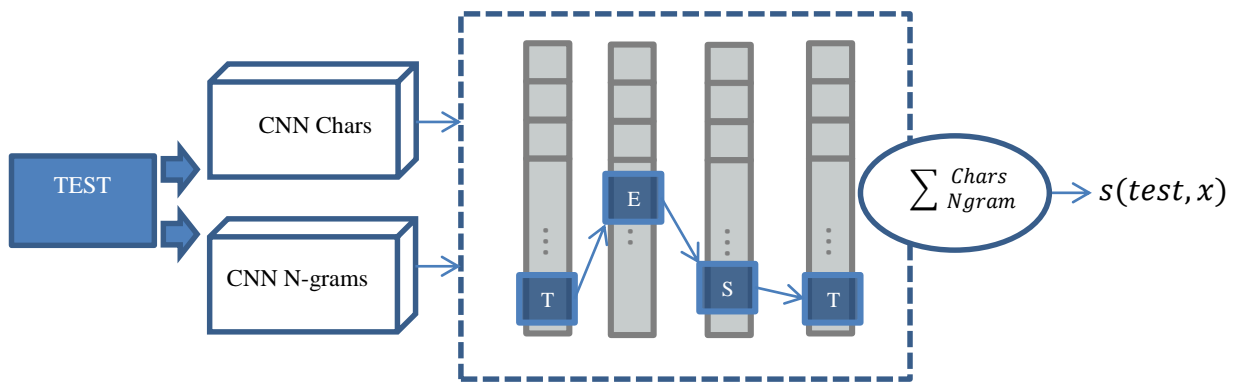

Figure 3:4 The joint model joins the character sequence and n-grams CNN models.

The joint model is CNN-based architecture that incorporates a conditional random field (CRF). The character sequence encoding model predicts the unaries for the CRF model and higher order terms are predicted by the n-gram encoding model.

The model does not include lexicon and free of constraints like maximum word length, but use statistical information from the n-gram model. The n-gram model makes use of the n-gram frequencies from a corpus and scales the gradient by the inverse of the frequency of the n-gram appearance.

## 3.5    Sequence-to-Sequence Learning with Neural Networks

The LSTM has been recently used for many computer vision and natural language processing problems. Mainly, the LSTMs are a good fit for NLP problems, where it performs very well in modeling sequences. A problem such as machine translation is a good and direct application for the LSTM [26]. The sequence to sequence learning with neural network [26] is the main motivation for our work. In this paper they teach LSTMs to map sequence of words in English to a sequence of words in French. The main contribution is how they handle sequence to sequence learning in arbitrary length. Another interesting paper uses LSTMs for computer vision problems [27]. Unsupervised learning of video representation using LSTMs encodes representation of video sequences into LSTM and decodes it using single or multiple LSTMs to perform tasks predicting the feature frames or reconstructing the input sequence.

The sequence to sequence learning with neural network is very relevant to our work where they use two multilayered Long Short-Term Memory (LSTM). One to map the input sequence to a fixed length vector and the other one to decode the target sequence from the vector.

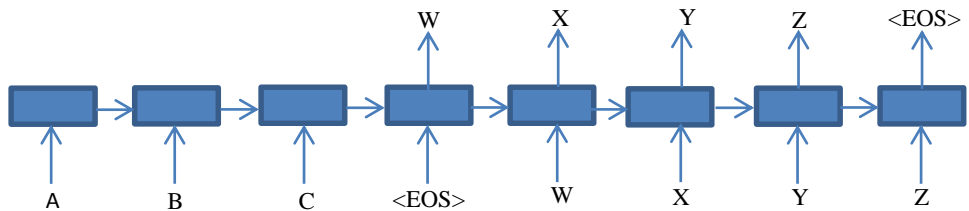

Figure 3:5 LSTM model reads an input sequence "ABC" and produces "WXYZ".

The LSTM ignores the output of $A, B, and\ C$ and as soon as the terminal character $< EOS >$ is read feeds the output character to the LSTM to generate the next one and terminate after producing the terminal character $< EOS >$.

# Chapter 4 Sequence-to-Sequence Learning for Unconstrained Scene Text Recognition

In this work we present a novel approach for unconstrained scene text recognition using convolutional neural networks (CNNs) and long short-term memory (LSTMs). We experimented with 18 different LSTM architectures in which we experiment fat, thin, deep, etc... See section 5.2.1 for in detail discussion. The aim for the 18 different architectures is to find the best performing LSTM architecture that corrects the CNN mistakes. The LSTM as sequence to sequence learning algorithm is able not only to learn the same length sequences but also can learn arbitrary length to length sequence as will be shown shortly.

## 4.1 Arbitrary Length Sequence to Sequence Modeling

In the sequence to sequence learning paper [26] the authors present an LSTM-based approach to translate from French to English. Instead we map the probability maps of the 37-classes $a - z, 0 - 9$, and the negative classes for all detected characters by the convolutional neural networks model for character sequence encoding [18] to a correct English word. One key difference between our task and the translation task is that we do not have grammar rules for mapping the character sequence probability maps to a correct English word. Our LSTM model does not have to learn implicitly grammar rules. So our problem is easier than the machine translation task.

The LSTM model, in sequence to sequence learning with neural networks [26], ignores the outputs of the input sequence's items till a terminal character is fed to the LSTM. Then it feeds the output of the terminal character $O_1$ to the LSTM as an input and gets $O_2$ as an output for $O_1$ which again feeds $O_2$ to the LSTM and gets $O_3$ etc... till the LSTM outputs a terminal character $< EOS >$ and stops detecting anything else. The order of words is given to the LSTM is important because in translation problems, the sentence structure changes from language to language according the grammar rules of each language.

We found from our experiments that the accuracy is improved if we start producing the output sequence after seeing the whole input sequence but it was slower in the experiment of extending CNN model with optimized LSTM architecture, section 5.3. We decided to use the maximum length trick and let the LSTM decides when it starts producing the output sequence regardless when the start point is, the LSTM has to produce an empty character at index 24 because the maximum length we use is 23 and comes from the maximum length of the CNN predictions.

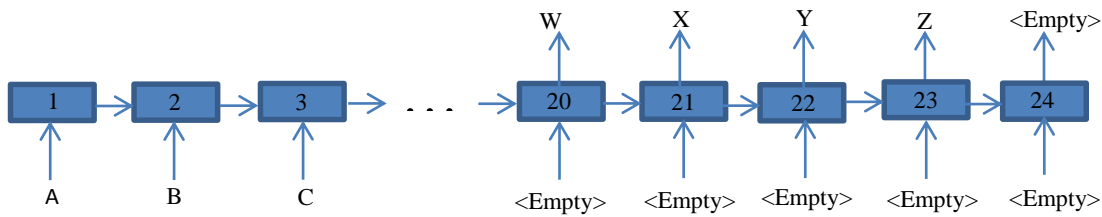

Figure 4:1 LSTM model reads an input sequence "ABC" and produces "WXYZ.

The empty characters come from padding all input sequences to the length of the CNN predictions (character sequence encoding model). Figure 4:1 shows an example for mapping $ABC$ to $WXYZ$ where the LSTM starts producing a sequence of length 4 characters at index 20 because it has to stop at index 23 and produces an empty character at index 24.

We use context scheme and right-justify the ground truth as shown in figure 4:1. We teach the LSTM to output empty characters at beginning and an empty character again "terminal character" to terminate the prediction process. The LSTM predicts characters, non-empty characters, after seeing enough probability maps characters to start the prediction process. In other words, the LSTM generates empty characters, a correct English word, from the LSTM point of view, and then another empty character as terminal as shown in figures 4:1 and 4:2.

Our aim is to teach the LSTM to map the false positives to empty characters, add the missing ones, replace the incorrect characters with the correct ones, and then output an empty "terminal" character. So, we handle the arbitrary length sequence to sequence in a different and more efficient way than what presented in sequence to sequence learning paper with neural networks [26].

We use only one LSTM model not two LSTMs. In [26] the first model is used for encoding the input sentence and the second one is for decoding the target sequence. Moreover in our model the LSTM can start producing characters before the input sequence is finished as shown in figure 4:2. So, it does not have to feed the predicted sequence to the LSTM. In our formulation, the LSTM learns when it starts producing characters and then producing an empty "terminal" character.

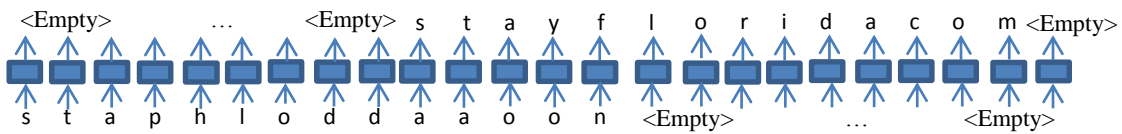

Figure 4:2. An example from the ICDAR test set.

The character sequence encoding model generated "staphlodaaoon" where the ground truth was "stayflorida.com" and our model corrected it to "stayfloridacom".

## 4.2   Training

We train the CNN and LSTM models using back propagation and stochastic gradient descent. We use the default learning rate= 0.0001 and momentum= 0.9 of lasagne library. CNN network is trained by back-propagating the gradients from each 23 softmax classifier to the base net, and LSTM using back-propagation through time. The maximum sequence length is 23 as presented in sections 3.5 and 4.1.

The CNN network training is performed solely on the synthetic dataset [18], while the LSTM due to memory constraints is trained on only 1.5M randomly sampled from the synthetic dataset besides the ICDAR 2013 training set. See the datasets description in appendix B.

The goal of the CNN network is to learn robust features and detect an initial character sequence the training of character sequence. A word $w$ of length $N$ is a sequence character $w = (c_1, \dots, c_{23})$ where $c_i \in C = \{1, 2, \dots, 37\} = \{0 - 9, \ a - z, \ \emptyset\}$ represents a character at position $i$ in the word $w$, where $\emptyset$ is an empty character.

For a given word $x$, we want to estimate $w^*$ that maximizes $P(w^*|x)$. Assuming the characters are independent we can predict $w^*$ as follow.

$$w^* = arg \ max_w P(w|x) = \arg \ max_{c_1, c2, \dots, c23} \prod_{i=1}^{23} P(c_i|g(x))$$

where $P(c_i|g(x))$ is given by the classifier for the $i^{th}$ character. Due to the independence assumption we can compute $w^*$ by taking the most probable character at each position $c_i^* = \arg \max_C P(c_i|g(x))$. The independence assumption would not let the network learns that 1g99 should be 1999 instead.

For the LSTM we teach the network to estimate $P(y_1, \dots, y_{23} \mid x_1, \dots, x_{23})$ where $(x, \dots, x_{23})$ is an input sequence and $(y_1, \dots, y_{23})$ its corresponding output sequence whose length 23 followed by an empty character. As shown in section 2.4.1 the output at time $t$ depends on the input of the all sequence till time $t - 1$

$$o_t = c + V s_t$$

where $s_t$ the state at time $t$ and implicitly encodes information about the whole sequence in the past $(x_t, x_{t-1}, \dots, x_2, x_1)$.

$$s_t = G_t(x_t, x_{t-1}, \dots, x_2, x_1)$$

## 4.3   Lasagne

We use lasagne library [28] for the LSTMs implementation, training, and testing. The library is underdevelopment and is written in python and Theano [28]. Lasagne supports GPU which makes the training time 10 to 50 times faster than training on a CPU. The library is open source making it possible to contribute or report bugs. The library documentation is well written and we found it very easy to install and understand. Theano functions are pretty different in terms of the syntax, input and out parameters. The functions are compiled as part of the symbolic graph which makes it hard to debug. The way you can debug it is you put the variable you would like to see its values in the output (the function's returned variables) and print them after the function finishes execution. The shared variables are the variables that you share theirs contents with the GPUs. So, we recommend putting only the part of your data set (current epoch) into these variables, otherwise your code may run out of GPU memory.

# Chapter 5    Experiments

In this section we shall present multiple experiments to show how LSTMs can improve the performance of a CNN and even can beat the state of the art Deep structured output learning for unconstrained text recognition [12], referred to as the joint model.

## 5.1    Extending CNN Model with LSTM for error correction

We use a pre-trained CNN model « detnet_layers.mat » [11] in this experiment. In the testing phase each image is scaled to 16 different scales and convolved with the convolutional neural network's filters to generate character maps. Then we use a maxout layer to choose for each pixel the character with highest probability of being certain character, and end up with a labels map. From the labels map we run a connected component algorithm [16] to combine the neighboring pixels with the same label together in the same component. Then we combine the character boxes $c_1$ and $c_2$ into a word box if the distance between the centers of $c_1$ and $c_2$ is less than minimum width between $c_1$ and $c_2$. Then, we use long short term memory to remove the false positive characters as shown in figure 5:1.

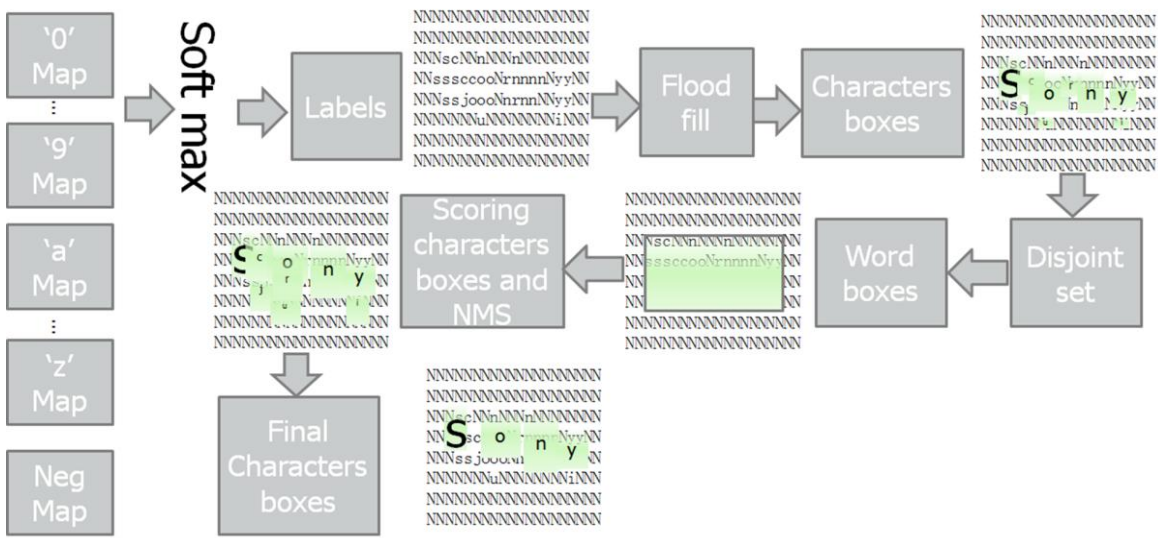

Figure 5:1 Shows how we move from the probability maps to the characters' bounding boxes.

We use two datasets for our experiments. The first one is the synthetic dataset [18] and the other one is ICDAR dataset. The synthetic dataset contains 90K English words; around 100 images for each word and ICDAR contains around 6K annotated words; 3564 for training and 1439 for testing. We use the synthetic dataset to enrich our training set. We choose images from the synthetic dataset to cover most of the words in the ICDAR dataset. Since the aim of the LSTM is to correct the CNN output, in this section we are going to present multiple experiments comparing the CNN performance versus that of the CNN+LSTM. For convenience we would call the latter as simply LSTM. For the CNN output, we either consider all produced character maps (referred to as CNN-Noisy), or use non-maxima suppression to filter the overlapping characters with overlap of at least 30% and keep the ones with high confidence (referred to as CNN-Clean) as shown in figure 5:1.

The LSTM model in this experiment uses context of length 3, which means it has to output at least 3 empty characters before producing the correct word and then another empty character as terminal as shown in figure 5:2.

In this experiment we use learning rate = 0.0001 and momentum= 0.9. We use stochastic gradient descent with patch size = 256. The LSTM converges after around 100 epochs on the validation set. The validation set consist of 800 images.

We shall present two evaluation techniques. The first one is a word-level and the second one is a character-level evaluation. For the word level evaluation we calculate the edit-distance between the model predictions and the ground truth words. For the word level we calculate the percentage of perfect match "PM" or zero distance between the model predictions and the ground truth, the percentage of words needing one operation (insertion, deletion, or swap) to match the model's predictions with the ground truth "ED-1", two "ED-2", three "ED-3", and more than three operations "ED>3".

| | PM | ED-1 | ED-2 | ED-3 | ED>3 |
|---|---|---|---|---|---|
| CNN-Clean | 15.8 | 21.6 | 18.6 | 12.2 | 31.8 |
| LSTM-Clean-03 | 37.7 | 15.0 | 12.7 | 10.1 | 24.5 |
| LSTM-Clean-10 | **41.1** | 13.2 | 10.4 | 10.5 | 24.8 |
| CNN-Noisy | 0.9 | 1.9 | 3.3 | 5.0 | **89.0** |
| LSTM-Noisy-03 | 37.1 | 13.7 | 11.7 | 11.3 | 26.2 |
| LSTM-Noisy-10 | 38.4 | 12.1 | 11.3 | 10.8 | 27.3 |

Table 5:1 shows the word level evaluation when we use context of three and ten.

In table 5:1 the LSTM shows significant improvement over the CNN, in perfect match column 37.7% for LSTM vs 15.8% for CNN more than 100% gain and 52.7% vs 37.4% for both perfect match and words needing one operation to match the ground truth around 40% gain for the LSTM.

| | Precision | Recall | F-measure |
|---|---|---|---|
| CNN-Clean | 0.584 | 0.646 | 0.613 |
| LSTM-Clean-03 | 0.656 | 0.650 | 0.653 |
| LSTM-Clean-10 | 0.665 | **0.667** | **0.666** |
| CNN-Noisy | 0.222 | **0.747** | 0.342 |
| LSTM-Noisy-03 | 0.665 | 0.630 | 0.647 |
| LSTM-Noisy-10 | **0.666** | 0.613 | 0.638 |

Table 5:2 shows character-level evaluation with context of 3.

The Longest Common Subsequence (LCS) between the predicted word and the ground truth is the true positives, the difference between the length of the ground truth and the LCS is the false negatives, and the difference between the length of the prediction and the LCS is the false positives. Precision and recall are calculated as follow.

$$Precision = \frac{true\ positive}{true\ positive\ +\ false\ positive}$$

$$Recall = \frac{true\ positive}{true\ positive\ +\ false\ negative}$$

Interestingly from table 5:2 we can notice that CNN-Noisy without non-maxima suppression has the highest recall and lowest precision which means it produces many false positives to detect around 75% of the all characters, but we lose around 10% of the all characters when we try to improve the precision either with non-maxima suppression or with an LSTM. The substantial improvement for the precision with LSTM and keeping the recall almost the same, confirms our argument that the LSTM can filter out the false positives and even can add true positives when missed by the CNN or non-maxima suppression. In terms of F-measure LSTM-Clean shows an improvement over CNN-Clean by 4% and almost 100% improvements for the LSTM-Noisy over CNN-Noisy.

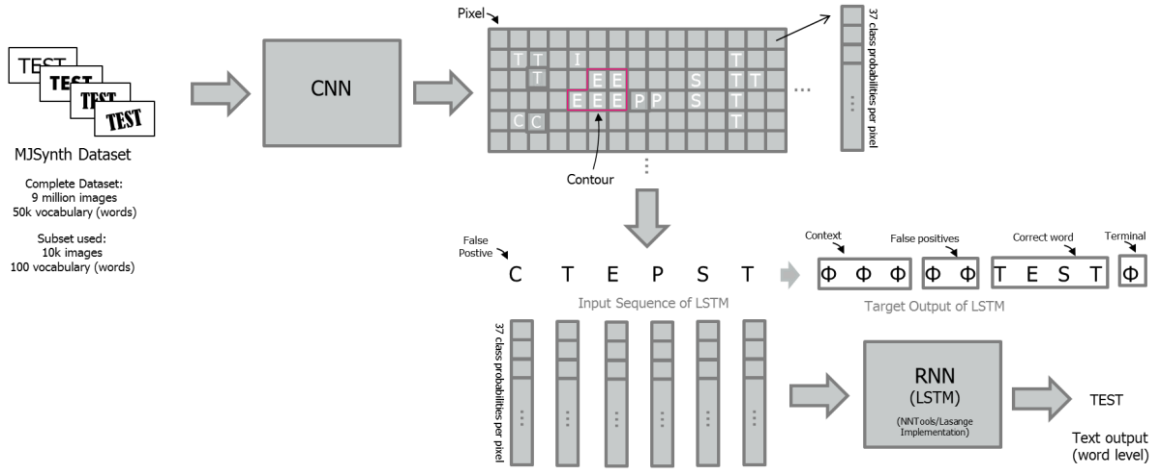

Figure 5:2 Shows how the LSTM corrects the CNN output from CTEPST to TEST

## 5.2    LSTM Architecture Experiments

In this experiment we use another CNN model that was published alongside with the paper "Synthetic Data and Artificial Neural networks for Natural Scene Text Recognition" [18]. For convenience in the remaining sections we will call it the CNN_Chars model. As mentioned in the previous experiment, we use the same learning rate, momentum, and patch size. We also consider both ICDAR and the synthetic datasets for training and ICDAR test set for testing. Also to allow for reasonable training times, around one day, we did not consider the whole synthetic dataset for training. We chose the images that cover most of the words in the training and test of ICDAR dataset. The aim of this experiment is to find out whether this approach would work or not and to pick the best performing LSTM architecture for the given task. In this experiment we tested 18 different LSTM models. Sometimes the architecture is deep and sometimes it is fat or thin. Moreover, we tried to intertwine dropout layers with different dropout rates. We compare the performance of the 18 models with the CNN_chars, and the joint (Deep Structured Output Learning for Unconstrained Text Recognition) [12] models. The joint model combines the CNN_chars output with another convolutional neural network based model for n-gram detection as explained in the background chapter. In the following section we shall present the different model architectures and how their performance compares to each other.

### 5.2.1  The models' architecture

The first model consists of an input layer, output dense layer followed by softmax for classification, and two bidirectional LSTM layers with number of nodes 102, and 156 for the first and second layers respectively as shown in figure 5:3.  Model no.1 is from the sample file of the Lasange library.

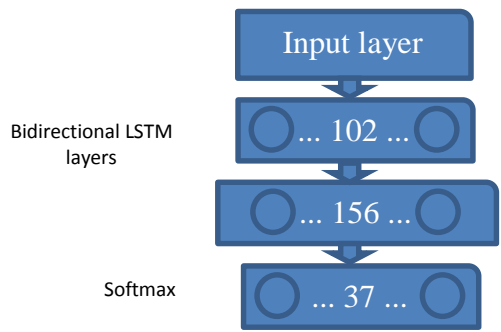

Figure 5:3 shows the LSTM architecture for model no. 1

The training for model no. 1 converged after nearly 100 epochs and the results were better than the CNN_chars model, but worse than the joint model as shown in the table 5:3.

|  | Accuracy |
|---|---|
| Model no. 1 | 80.0% |
| Joint | 81.8% |
| CNN_Chars | 79.5% |

Table 5:3

The results, as initial results, are good and encouraged us to try other architectures, like adding more bidirectional LSTM layers (BLSTM layers), changing the number of nodes per layer for the BLSTM layers, and adding dropout layers.

The second and third architectures, figure 5:4, just added one more bidirectional LSTM layer. The second, model no. 2, architecture is fatter. In other words, the number of nodes per layer is bigger than the third one, model no. 3.

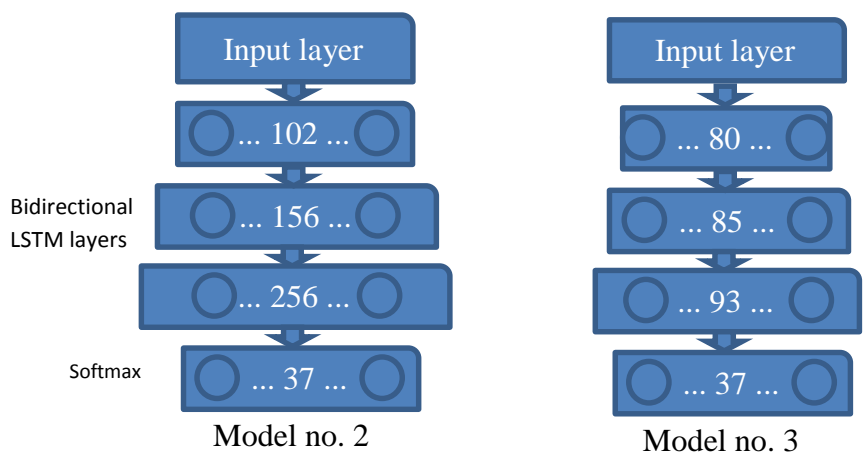

Figure 5:4: shows the LSTM architecture for model no. 2 and model no. 3

According to table 5:4 Model no. 3 performs better than model no. 2 by 0.5% but model no. 2 didn't improve over model no. 1.

The picture is not yet clear, going deep didn't improve, going thin improved by 0.5% when comparing the performance of model no. 3 with model no. 2 and going deeper and thinner improved by 0.5 when comparing the performance of model no. 3 with model no. 1.

| | Accuracy |
|---|---|
| Model no. 1 | 80.0% |
| Model no. 2 | 80.0% |
| Model no. 3 | 80.5% |
| Joint | **81.8%** |
| CNN_Chars | 79.5% |

Table 5:4

In model number four we added one more bidirectional LSTM layer and all of them of the same number of units, 200 units as shown in figure 5:5. We were expecting reducing number of model parameters would affect the accuracy in a bad way. Surprisingly, the accuracy is improved by 1.4% compared to model no. 1. This model was a good motivation to test even more deep architectures, although we have shown that a model of 5 BLSTM layers is too deep and resulted in lower accuracy. Model no. 5 consists of one input, one output followed by softmax, and 5 BLSTM layers each of 150 units. The accuracy is 81.1% for model no. 5 compared to 81.4% to model no 4.

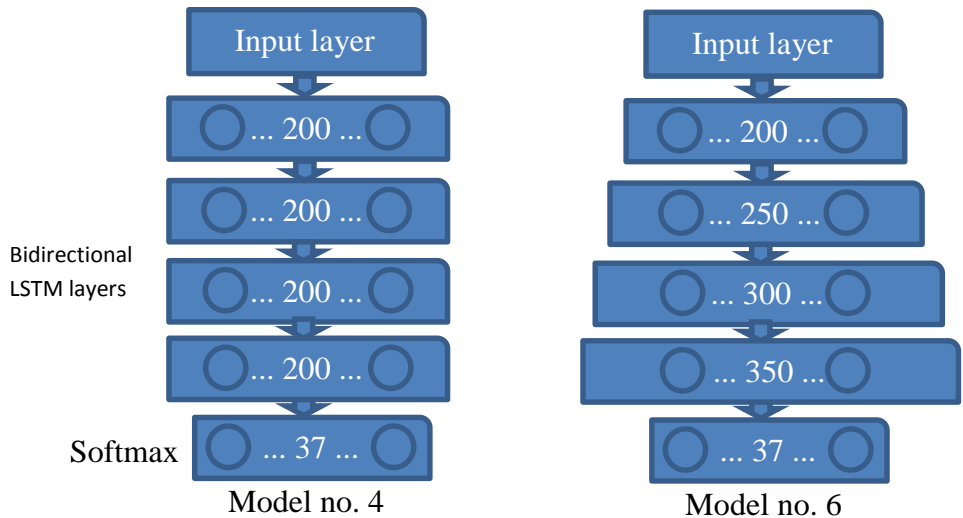

Figure 5:5 shows the LSTM architecture for model no. 4 and model no. 6

Model no. 6 is 4-BLSTM layers like model no. 4 but with different number of nodes per layer. Number of nodes in each BLSTM layer is 200, 250, 300, and 350. The accuracy of model no. 6 is improved by 0.7% compared to model no. 4 and the first to outperform the joint model as shown in table 5:5.

| | Accuracy |
|---|---|
| Model no. 1 | 80.0% |
| Model no. 4 | 81.4% |
| Model no. 6 | **82.1%** |
| Joint | 81.8% |
| CNN_Chars | 79.5% |

Table 5:5

It was quite interesting to add one more BLSTM layer over model no. 4 (model no. 7) with 200 nodes and one more layer over model no. 6 (model no. 8) with 400 nodes. The results of model no. 7 did not improve over model no. 4. However it was worse by 0.1% than model no. 4 and the accuracy of model no. 8 was worse than model no. 6 by 0.7%.

Model no. 9 was a 4-BLSTM-layers model like model no. 6 in terms of number of BLSTM layers but more number of nodes per layer (fatter). Number of nodes in each BLSTM layer was 300, 350, 400, 450, and 500. Although model no. 9 contains more parameters than model no. 6, the accuracy of model no. 9 was worse than model no. 6 by 0.7%. Moreover, models no. 10 and 11 of 6-BLSTM layers did not improve over model no. 6 the 4-BLSTM layers model. The accuracy of model no. 10 even is much worse than the accuracy of model no. 6. The message was the deeper we go, not necessarily the better performance is achieved as shown in table 5:6.

|  | Accuracy |
| --- | --- |
| Model no. 10 | 80.2% |
| Model no. 11 | 82.0% |
| Model no. 6 | **82.1%** |
| Joint | 81.8% |
| CNN_Chars | 79.5% |

Table 5:6

In models no. 12 and 13 we tried the reverse shape of model no. 6 but both did not improve the accuracy as show in table 5:7. Model no. 12 is a 4-BLSTM layers model with number of units 350, 300, 250, and 200, and model no. 13 is a 5-BLSTM layers model with number of units 500, 450, 300, 250, and 200.

|  | Accuracy |
| --- | --- |
| Model no. 12 | 81.8% |
| Model no. 13 | 82.1% |
| Model no. 6 | **82.1%** |
| Joint | 81.8% |
| CNN_Chars | 79.5% |

Table 5:7

We added dropout layers starting from model no. 14 and it improved the accuracy dramatically as shown in table 5:8.

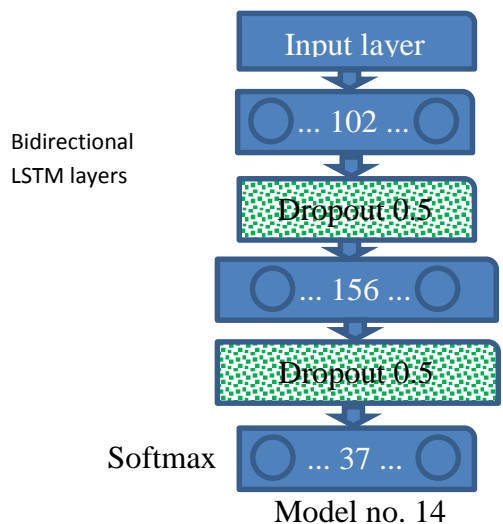

Model no. 14

Figure 5:6 shows the LSTM architecture for model no. 14

Model no. 14 is the same as model no. 1 but with dropout layers intertwined with dropout rate 0.5 as shown in figure 5:6. The model has converged after around 200 epochs due to the dropout layer.

|  | Accuracy |
|---|---|
| Model no. 1 | 82.0% |
| Model no. 14 | **83.0%** |
| Joint | 81.8% |
| CNN_Chars | 79.5% |

Table 5:8

Model no. 15 is the same as model no. 12 but with dropout layers are added after each BLSTM layer except the first one, and the dropout rate is 0.5. Adding one more BLSTM layer, model no. 16, on top of model no. 15 didn't change the accuracy as in table 5:9.

The best performing model among all the 18 experimented architectures is model no. 17. It consists of 4-BLSTM layers. Each BLSTM layer is followed by a dropout layer with dropout rate 0.5, input layer, and output dense layer followed by softmax for classification as shown in figure 5:7. See Appendix A.

|  | Accuracy |
|---|---|
| Model no. 12 | 81.8% |
| Model no. 15 | **83.5%** |
| Model no. 16 | **83.5%** |
| Joint | 81.8% |
| CNN_Chars | 79.5% |

Table 5:9

Moreover we changed the dropout rate to 0.3 instead of 0.5 for model no. 17 but the accuracy went down 1.0% worse than model no. 17 as shown in table 5:10.

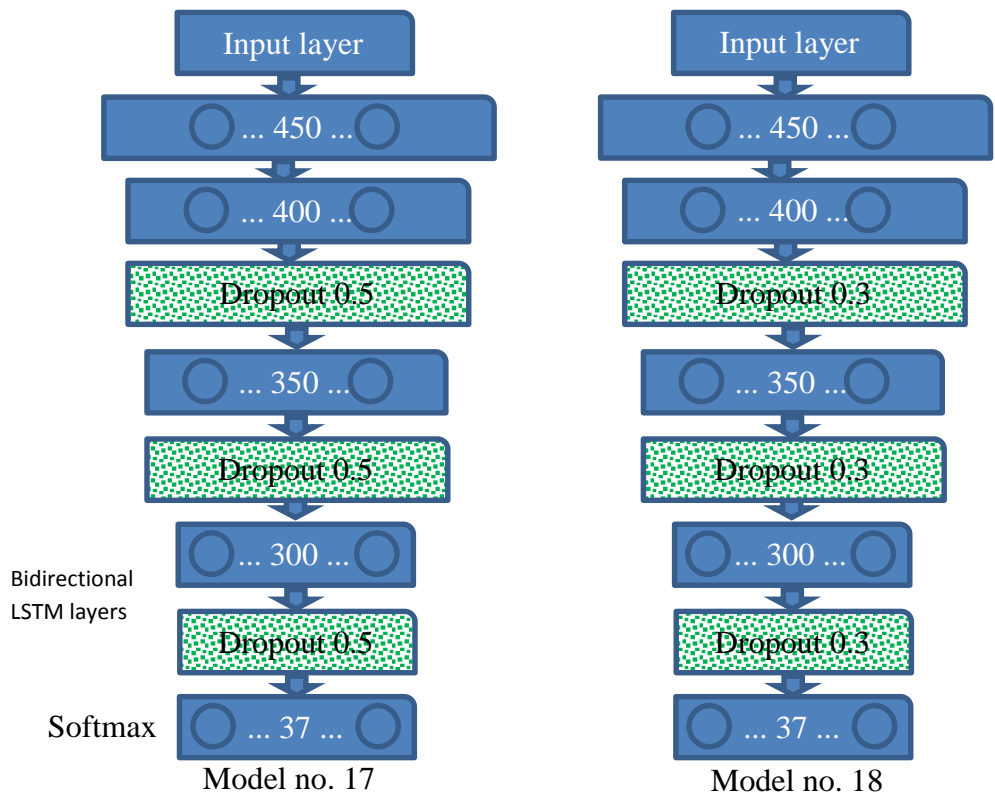

Figure 5:7 shows the LSTM architecture for model no. 14

Due to the good performance of model no. 17 as show in table 5:10, we choose it for our big experiments as will be explained shortly next section.

|  | Accuracy |
| --- | --- |
| Model no. 17 | **84.3%** |
| Model no. 18 | 83.1% |
| Joint | 81.8% |
| CNN_Chars | 79.5% |

Table 5:10

We experimented with architectures of the LSTM networks, to see how the factor such number of layers, nodes per layer etc... would affect the performance. These experiments were not comprehensive, but indicate which features generally improve performance. From our limited set of experiments we found out that model no 17 is the best performing model so we use it for our large experiments as shown in table 5:11.

|  | Accuracy | # BLSTM Layers | # nodes in each layer | Dropout P. |
| --- | --- | --- | --- | --- |
| Model no. 1 | 80.0% | 2 | 102, 156 | - |
| Model no. 2 | 80.0% | 3 | 102, 156, 256 | - |
| Model no. 3 | 80.5% | 3 | 80, 85, 93 | - |
| Model no. 4 | 81.4% | 4 | 200, 200, 200, 200 | - |
| Model no. 5 | 81.1% | 5 | 150, 150, 150, 150, 150 | - |
| Model no. 6 | 82.1% | 6 | 200, 250, 300, 350 | - |
| Model no. 7 | 81.3% | 5 | 200, 200, 200, 200, 200 | - |
| Model no. 8 | 81.4% | 5 | 200, 250, 300, 350, 400 | - |
| Model no. 9 | 81.4% | 5 | 300, 350, 400, 450, 500 | - |
| Model no. 10 | 80.2% | 6 | 100, 120, 140, 160, 170, 180 | - |
| Model no. 11 | 82.0% | 6 | 300, 320, 340, 360, 370, 380 | - |
| Model no. 12 | 81.8% | 4 | 350, 300, 250, 200 | - |
| Model no. 13 | 82.1% | 5 | 500, 450, 300, 250, 200 | - |
| Model no. 14 | 83.0% | 2 | 102, 156 | 0.5 |
| Model no. 15 | 83.5% | 4 | 350, 300, 250, 200 | 0.5 |
| Model no. 16 | 83.5% | 5 | 350, 300, 250, 200, 150 | 0.5 |
| Model no. 17 | **84.3%** | 4 | 450, 400, 350, 300 | 0.5 |
| Model no. 18 | 83.1% | 4 | 350, 300, 250, 200 | 0.3 |
| Joint | 81.8% | - | - | - |
| CNN_Chars | 79.5% | - | - | - |

Table 5:11 Models descriptions

## 5.3  Extending CNN Model with optimized LSTM Architecture for error correction

In this experiment we make use of a large amount of images that was chosen randomly from the synthetic dataset to enrich the ICDAR training set. We picked around 1.5M million images covering around 80K words, around 20 synthetic images for each word. The training phase takes around 4 days because we use a big dataset for this experiment. We use the architecture of model no. 17 and the same parameters for learning rate = 0.0001, momentum = 0.9 and patch size = 256 from the previous experiment. The model has converged after around 200epochs due to the dropout layer.

The results show that we have a little improvement over the state of the art and significant improvement over the CNN_Chars model [18] as shown in table 5:12. Since our approach employs the CNN_Chars output and feeds it to a Long Short Term Memory network, it should be named CNN+LSTM. For convenience we will call it LSTM instead of CNN+LSTM.

|  | Accuracy |
|---|---|
| LSTM | **82.0%** |
| Joint | 81.8% |
| CNN_Chars | 79.5% |

Table 5:12

We printed out the words that the LSTM predicts correctly and the CNN_Chars corrupts vs the words that the LSTM corrupts and the CNN_Chars predicts correctly and we found some interesting phenomena. Starting with words needing one operation to match the ground truth.

| Ground Truth | CNN_Chars | LSTM |
|---|---|---|
| Plugin | Plugin | Plugin |
| Through | through | Through |
| Customers | customerss | Customers |
| Zpa | Zppa | Zpa |
| Sleeps | ispleeps | Sleeps |
| 200 | 200t | 200 |
| Today | Today | Today |
| Was | Wasi | Was |
| Now | Nowl | Now |
| Com | Icom | Com |
| The | Thel | The |
| Save | Saver | Save |
| 50 | 507 | 50 |
| Per | Perr | Per |
| Time | Timei | Time |
| Question | ouestion | Question |
| 860 | 360 | 860 |
| 4gp | Agp | 4gp |
| Digikey | Digiley | Digkey |
| Click | Cuick | Click |
| Go | 90 | Go |
| Ieee | Jeee | Ieee |
| Multicomp | multicmmp | Multicomp |

Table 5:13 The LSTM is correcting words needing one operation to match with the ground truth which are produced by the CNN.

From table 5:13 we can see that the LSTM corrected all type of 1-operation errors, like deleting an extra character, adding a missing character, and changing an incorrect one. Deleting an extra character like in (plugin, through, zpa, sleeps, 200, today, was, now, the, save, 50, per and time), adding a missing character like in (question), and changing an incorrect character like in (860, 4gp, digikey, click, go, ieee, and multicomp).

Words 860, go, ieee and 4gp we can see that '8' visually looks like '3', 'g' looks like '9', 'i' looks like 'j', and '4' looks like 'a'. So, we can argue that those words are corrected due to this visual similarity. Words digikey, click, and multicomp the visual similarity doesn't exist anymore because 'm' does not look like 'o', 'u' does not look like 'l', and 'l' does not look like 'k'. So, may be the LSTM relied on the n-gram statistics from the training set and contextual information from the surrounding characters. The same argument of n-gram statistics and the contextual information from the surrounding characters applies to the cases of deleting an extra character and adding a missing one.

Interestingly LSTM can even correct words needing more than one operation to match the ground truth as shown in table 5:14. For example 256bit is detected as zsobit by the CNN_chars model. The LSTM was able to correct zso to 256.

| Ground truth | CNN_chars | LSTM |
|---|---|---|
| Sleeps | Islepps | Sleeps |
| 10 | 1010 | 10 |
| 1118th | Ilisth | 1118th |
| 256bit | Zsobiit | 256bit |
| 156 | 1651 | 156 |
| Must | Ko | Must |
| 50 | so7 | 50 |
| Cars | Tcass | Cars |
| 1999 | 19g6 | 1999 |
| 10 | 1000 | 10 |
| Stayfloridacom | Staphloddaaoon | Stayfloridacom |
| 1417 | 1717 | 1417 |
| Mhealth | Mhlaath | Mhealth |

Table 5:14 The LSTM is correcting words needing more than one operation to match with the ground truth which are produced by the CNN.

LSTM not only corrects words need one operation to match with ground truth, but also words needing more than one operation and even a combination of adding, removing, and swapping characters. The table 5:15 shows the words that the LSTM model corrupted them.

| Ground truth | CNN_chars | LSTM |
|---|---|---|
| thank | thank | Hhank |
| us | Us | Uss |
| 360 | 360 | 160 |
| sailings | sailings | Saaliiggi |
| open | open | Opent |
| month | month | Monthi |
| problem | Problem | Probbeem |
| cashback | Cashback | Hashback |
| berlin | Berlin | Berline |
| 25th | 25th | 25t |
| best | Best | Bett |
| question | Question | Ouestion |

Table 5:15 The LSTM is corrupting words which are produced correctly by the CNN.

## 5.4 Generalization Experiment

In this experiment we show that the LSTM can correct words that have never been seen in the training set. We use the same training and test set as described in experiment 3 but we remove from the 1.5M synthetic images all occurrences of the words that exist in the ICDAR test set. The LSTM has corrected 44 words which we present in table 5:17 and spoiled 34 words presented in table 5.18. The accuracy was still better than the CNN_Chars model but worse than the joint model as in table 5:16. The aim of this experiment is to check whether the LSTM can detect unseen words or not.

|          | Accuracy |
|----------|----------|
| LSTM     | 80.2%    |
| Joint    | **81.8%** |
| CNN_Chars | 97.5%   |

Table 5:16 generalization test.

| Ground truth    | CNN_chars       | LSTM           |
|-----------------|-----------------|----------------|
| plugin          | plugiin         | Plugin         |
| Anfield         | anfilld         | anfield        |
| sleeps          | islepps         | Sleeps         |
| Sleeps          | islepps         | Sleeps         |
| 2009            | 2008            | 2009           |
| through         | throughh        | through        |
| customers       | customerss      | customers      |
| mortgages       | mortgagess      | mortgages      |
| Clinic          | cciinc          | Clinic         |
| sleeps          | isleeps         | Sleeps         |
| Sleeps          | isleeps         | Sleeps         |
| 10              | 1010            | 10             |
| include         | includes        | include        |
| 200             | 200t            | 200            |
| today           | today           | Today          |
| 4gb             | agb             | 4gb            |
| save            | saver           | Save           |
| digikey         | digiley         | digikey        |
| fast            | fasts           | Fast           |
| click           | cuick           | Click          |
| must            | ko              | Must           |
| go              | 9o              | Go             |
| 50              | 507             | 50             |
| go              | 9o              | Go             |
| 50              | 507             | 50             |
| must            | ko              | Must           |
| go              | 9o              | Go             |
| 50              | so7             | 50             |
| ieee            | jeee            | Ieee           |
| per             | perr            | Per            |
| per             | perr            | Per            |
| free            | frie            | Free           |
| multicomp       | multicmmp       | multicomp      |
| discount        | discountt       | discount       |
| 1999            | 19g6            | 1999           |
| per             | perr            | Per            |
| the             | thee            | The            |
| highspeed       | highaspeed      | highspeed      |
| code            | coder           | Code           |
| question        | ouestion        | question       |
| com             | icom            | Com            |
| stayfloridacom  | staphloddaaoon  | stayfloridacom |
| now             | nowl            | Now            |
| it              | ntt             | It             |

Table 5:17 Shows that the LSTM can detect unseen words.

| Ground truth | CNN | LSTM |
|---|---|---|
| thank | thank | Thnnk |
| 2010 | 2010 | 2116 |
| save | save | Save |
| sailings | sailings | Sailiigsi |
| 236 | 236 | 228 |
| 02110 | 02110 | 60110 |
| 617 | 617 | 811 |
| 617 | 617 | 811 |
| 507 | 507 | 70 |
| amazoncouk | amazoncouk | amazonokuu |
| month | month | Month |
| firewall | firewall | Firewalls |
| all | all | Ail |
| auto | auto | Atto |
| easter | easter | Eavter |
| 29th | 29th | 42th |
| cashback | cashback | Rashback |
| laptops | laptops | Laptors |
| mykonos | mykonos | Mykonosi |
| 0800 | 0800 | 6600 |
| 358 | 358 | 668 |
| 1229 | 1229 | 1222 |
| berlin | berlin | Berlinc |
| 25th | 25th | 2277 |
| just | just | Jjsst |
| best | best | Bett |
| paypal | paypal | Aypail |
| groove10 | groove10 | Grooveio |
| siggraph2010 | siggraph2010 | siggraphhelo |
| ieee | ieee | leel |
| ieee | ieee | leeel |
| mwc | mwc | Mmc |
| 69 | 69 | 89 |
| mwc | mwc | Mmc |

Table 5:18 The LSTM is corrupting words which are produced correctly by the CNN.

# Chapter 6   Traffic monitoring

Autonomous navigation is an interesting application scenario where unconstrained scene text recognition plays a vital role in reading road signs and ads. In this chapter, we discuss two sub-problems in the domain of traffic monitoring: 1) recognizing license plate numbers on the road to help identifying speeding cars for instance and 2) counting the number of cars for flow control scenarios.

## 6.1   Recognizing license plate numbers

Scene text recognition can help in reading car plates. We feed the car boxes to the CNN and get the characters' response maps. Then use non-maxima suppression and disjoint set to find the licenses plate bounding box as explained in figure 6.1. We use the same algorithm from section 5.1 to form the bounding around the plate's text.

We have applied the proposed methods using CNN+LSTM to a collection of real-world images taken from a highway scene. Figure 6.2 presents a sample of our recognition results. The sample GOE 361 is recognized incorrectly due to the visual similarity between 'G' and 'C'.

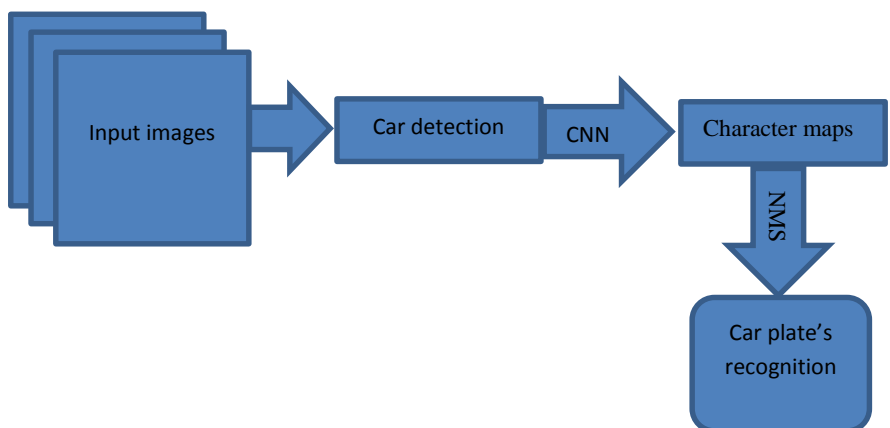

Figure 6:1 The car plate's recognition block diagram.

A necessary step in recognizing license plate numbers is to first localize cars as in figure 6.1. Hence we discuss next a novel algorithm for vehicles detection and tracking.

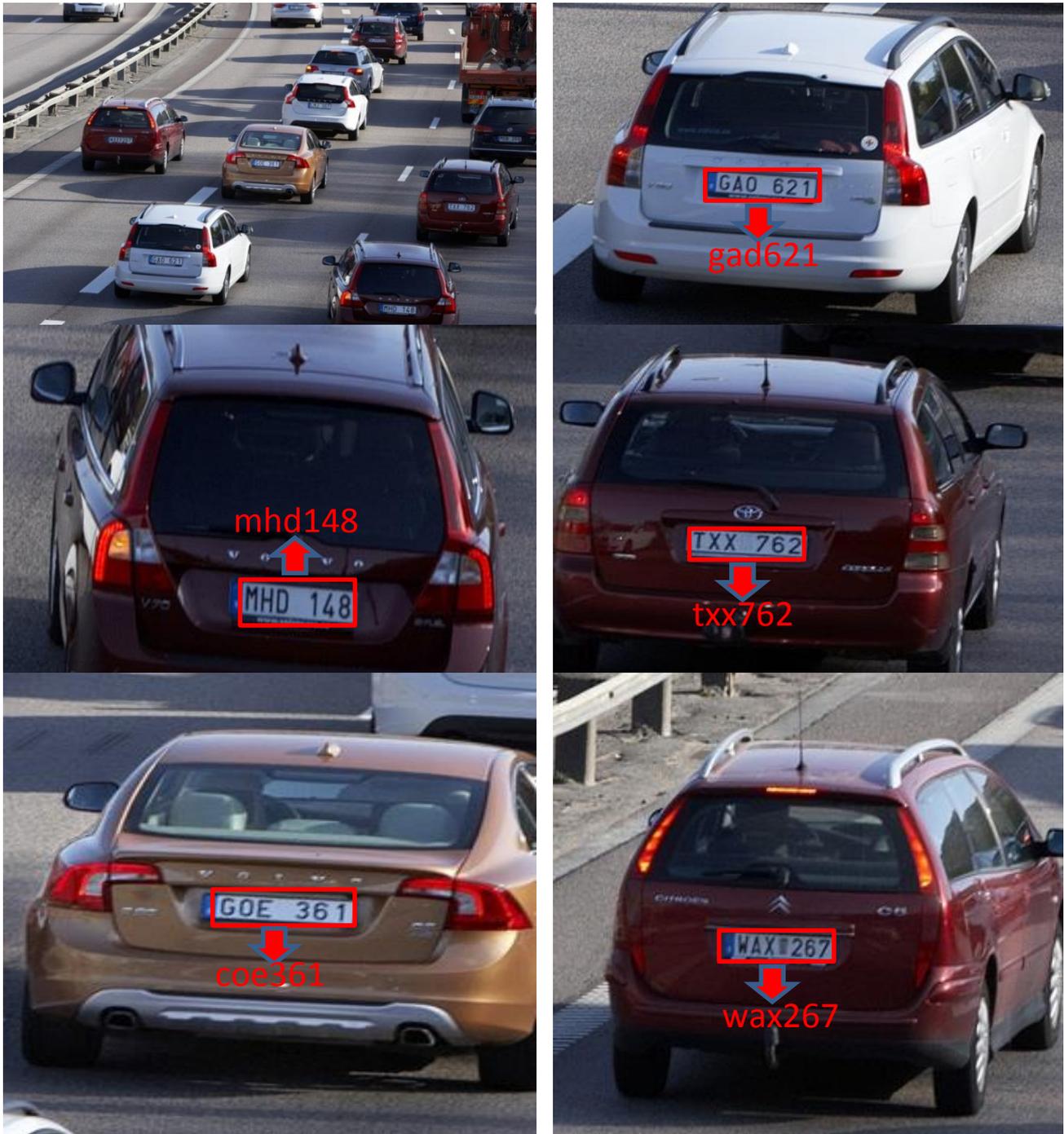

Figure 6:2 shows an example for car detection and car plate recognition

## 6.2    Counting and tracking vehicles

The second sub-problem we tackle in the domain of traffic monitoring is that of counting vehicles. Counting and tracking vehicles in very crowded scenes can help monitoring roads and choosing the path that takes the least amount of time which is not necessarily the shortest path in terms of length.

Counting and tracking vehicles in very crowded scenes is a very challenging task. In this task we do not require discipline conditions to deal with. We present a novel algorithm for automatically counting the number of moving vehicles in regular and very crowded scenes under different conditions. First we extract interest points and calculate trajectories independently. Then cluster the interest point into initial clusters based on proposed mathematical relations. We estimate number of moving vehicles by grouping the initial clusters based on a new adaptive background construction method, maximum sub-rectangle sum algorithm and disjoint set data structure. We have applied our algorithm on a collected dataset representing very crowded traffic scenes, where it shows an excellent high accurate performance. In addition, it has low storage and computational requirements, which promotes it for real time applications.

## 6.2.1 Extracting suitable corner points and their trajectories

The motion parameters of moving points in a sequence of images can be computed by identifying pairs of points that correspond to each other's in two images taken at time $t$ and $t + \delta t$.. It is not a good idea to match pixels with each other, due to illumination and lighting change between frames. We use corner points instead. A good property of corner points is that they can be precisely identified in an image rather than pixels' value in flat zones. Corner points [34] are points with high spatial gradient and high curvature. So the motion vectors associated with corner points are more reliable than raw pixels. We calculate the corner points for the foreground only where the foreground here are the moving objects. The difference between two consecutive frames will result in positive or negative values at the moving pixels and zeros everywhere else. Then we apply morphological operations to remove noisy regions. The filtered foreground pixels are grouped into regions employing connected component labeling algorithm. Corner point's detector is applied on each region which assures that each corner is a moving pixel. This pre-processing step will let us avoid calculating corner points for the whole image and facilitate clustering as well. After calculating corner points, optical flow, e.g Pyramid lucas kanade optical flow method [35], is applied on these points in frame $t$ to get the corresponding points in frame t + δt. Points are tracked over sequence of images using optical flow to calculate trajectories.

## 6.2.2 Initial Clustering

Clustering is an important step of our algorithm where trajectories that belong to the same object should be in the same cluster. The input of this step is the vector $Tr = \{Tr_1, Tr_2, \ldots, Tr_L\}$ Where $Tr_j$ is the $j^{th}$ trajectory and $L$ is the number of trajectories. A single trajectory $Tr_j$ is often represented as a path, which consists of sequence of points. It can be written as $Tr_j = \{P_j^1, P_j^2, P_j^3, \ldots, P_j^N\}$ where $N$ is the number of points in a trajectory. Given such input data, the goal is to produce a set of clusters $Clust = \{C_1, C_2, \ldots, C_M\}$ where $M$ is number of moving objects and each cluster consists of the same object's trajectories. We use the trajectories' angles, distance between trajectories, and parallel similarity [38] as parameters to determine whether two trajectories belong to the same cluster or not.

## 6.2.3 Adaptive Background Construction

We need to build a model for background to make detecting foreground's object an easy task.. Since background changes with time due to illumination and background variation, in this section we will describe an adaptive background construction method. We claim that the occurrence of background pixels values in non-moving parts is more frequent than any other value. We use a global 2D-array BK with the same size of the input video frame size representing the background model. We transform the input image from RGB to HSV color system. The background model BK consists of a pair of values $H$(the Hue of the pixel color) and $C$(counter representing the confidence of the corresponding hue value $H$). H and C for each background element will be updated as follow.

- The initial values of $H$ and $C$ are, $C = 0$ and $H = 0$
- For each pixel of the non-moving parts.
    - If $C = 0$ then $H$ = hue of the current pixel color value, and increase $C$ by 1.
    - If $C > 0$ and $H \neq$ hue of the current pixel color value, decrease $C$ by 1.
    - If $0 < C < C_{max}$ and $H$ = hue of the current pixel color value, increase $C$ by 1 and set $H$ = Hue of current frame, where $C_{max}$ is a constant representing the maximum confidence.

These steps are applied on all frames to detect light (day/night) and any other changes in the background model.

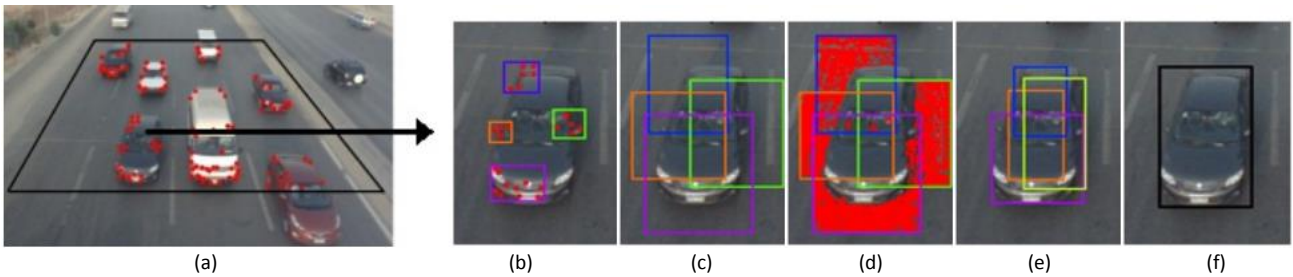

Figure 6:3. (a) Extracting suitable corner points, (b) Clusters obtained from the initial clustering step, (c) Rectangles Si around the initial clusters $C_i$, (d) Red points represent the negative score, (e) The subrectangles that have maximum sum and (f) Bounding box that merges all intersecting rectangles

## 6.2.4 Extracting bounding boxes around vehicles

For each cluster $C_i$ obtained from section 6.2.2 as shown in figure 6.3(b) an initial rectangle $I$ with size $h * w$ will be generated where the center of $I$ is the same as the center of $C_i$ as shown in figure 6.3(c). Each element of $I$ takes a negative score if the corresponding pixel is background as indicated by BK and positive score otherwise, as shown in figure 6.3(d). We aim to obtain the sub-rectangle $S_{max}$ of $I$ that covers the largest part of the moving object as shown in figure 6.3(e). It is worth to note that there are $h^2 * w^2$ possibilities to obtain a sub-rectangle inside $I$. Consider a matrix $S$, having $h$ rows and $w$ columns whose elements $s(i, j)$ retain the sum of the elements of the sub-rectangle $(1, 1, i, j)$ where $i = 1$ to $h$ and $j = 1$ to $w$. Using the matrix $S$ we are able to compute the sum of the elements of the current sub-rectangle $(i, j, k, l)$ in order of magnitude one $O(1)$. The Algorithm that calculates the matrix with the maximum sum is based on the Maximum Interval problem which is the one-dimensional array version of the maximum sub-rectangle problem. Using dynamic programming we can solve the Maximum Interval problem in order of $n$ where $n$ is number of elements in the interval, so, the same algorithm is applied for solving the maximum sub-rectangle problem in order of $h^2 * w$ time complexity $O(h^2 * w)$. Finally, the bounding box that bounds the vehicle shown in figure 6.3(f) is obtained by joining all sub-rectangles $S_i$ which are representing all clusters contributing in this moving object. It is supposed that the rectangles that share the same vehicle have a common intersection area. Those rectangles are joined using a disjoint set data structure [36] to obtain exactly one rectangle covering the vehicle. The complexity of joining those rectangles is $O(n)$ where $n$ is number of rectangles.

## 6.3    Experimental results and analysis

We have conducted three experiments on our newly collected dataset [37] which is presenting vehicles in very crowded scenes taken in different places and at different times. The dataset consists of three videos, NUvideoI, NUvideoII and NUvideoIII. In this dataset vehicles do not follow any lanes or fixed pattern. NUvideoI was recorded on a highway road during day time as shown in figure 6:2(a). Camera was installed on a six meters high. Number of vehicles passed through a predefined virtual zone was 3337 in 45 minutes. NUvideoII and NUvideoIII were recorded on ring road at two different times with different shadow conditions as shown in figure 6:2(b) and 6:2(c). Camera was installed on an eight meters high. Number of vehicles in ten minutes was 1141 and 1168, respectively. The first experiment was applied on NUvideoI where camera was attached with panda board 1M RAM and CPU Dual-core ARM Cortext m A9 MP with Symmetric Multiprocessing (SMP) at up to 1.2 GHz. The algorithm counted 3366 vehicles and recorded the time stamp of the entrance and exit of each vehicle which passed through the virtual zone and achieved an accuracy of 99.2% with a real time computation.

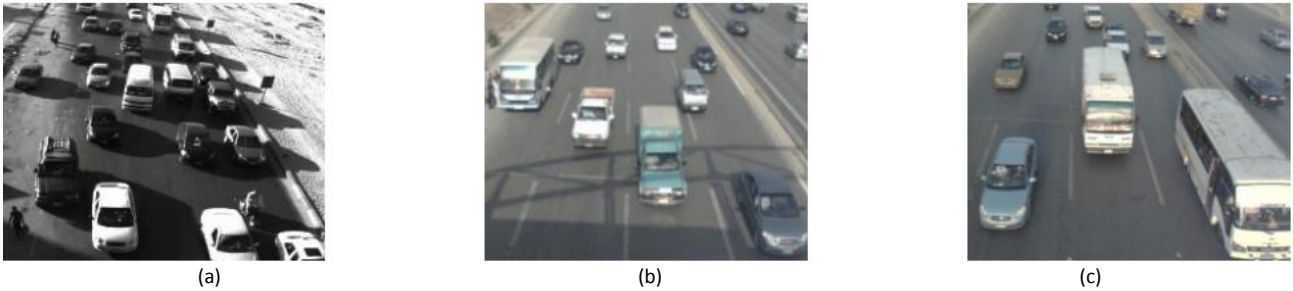

(a)    (b)    (c)

Figure 6:4.  (a) NUvideoI, (b) NUvideoII and (c) NUvideoIII dataset samples

The second and third experiments were applied on NUvideoII and NUvideoIII. These experiments were run on an Intel core(TM) 2 Duo T6500 2.1GHz PC with 4GB RAM. The algorithm counted 1151 and 1196 vehicles which achieve an accuracy of 99.13% and 97.60%, respectively. The average process time in our experiment was equal to 0.1 seconds per 5 frames which promotes our algorithm for real time applications. It is concluded that the algorithm count and track vehicles in real time with limited computational resources. Practical deployments shown in table 6:1 confirm these excellent properties.

| Video name | True no. vehicles | Estimated no. vehicles | Accuracy |
|---|---|---|---|
| NU video I | 3337 | 3366 | 99.2% |
| NU video II | 1141 | 1151 | 99.1% |
| NU Video III | 1168 | 1196 | 97.6% |

Table 6:1. Results of our method applied on our collected videos

# Chapter 7    Conclusion

## 7.1    Summary

In this thesis we presented a novel approach for unconstrained scene text recognition. We implemented a DNN approach incorporating convolutional neural networks with long short term memory. We showed that the LSTM is able to make use of contextual information and correct the output of the CNNs. Also we introduced a new way to handle arbitrary length sequence to sequence learning using long short term memory. Moreover we showed an application for the scene text recognition and presented a novel algorithm for vehicle counting and tracking.

## 7.2    Achieved results

We proposed an LSTM and CNN-based approach that achieve state of the art results 82% as shown in section 5.3, while being faster than the previous state of the art [12]. We also showed that the LSTM can generalize and predict words which were never seen in the training set. Moreover, we showed how factors such as number of layers, nodes per layer, and the dropout layers would affect the LSTM performance. Also, the performance of our algorithm for vehicle counting and tracking

## 7.3    Future development

The joint model shows a significant improvement over the character sequence encoding model (CNN Chars). So, In the future we plan to incorporate the output of the CNN+LSTM with the output of the CNN n-gram network from [18] (joint-LSTM). We expect to get better results than the joint model because the n-gram CNN-based model detects contextual information from the images directly and our model learns this contextual information from the English words and corrects some of the CNN Chars model mistakes.

The joint-LSTM is the model that combines CNN Chars + LSTM and CNN n-gram model, and the joint-CNN is the one that combines CNN Chars and CNN n-gram.

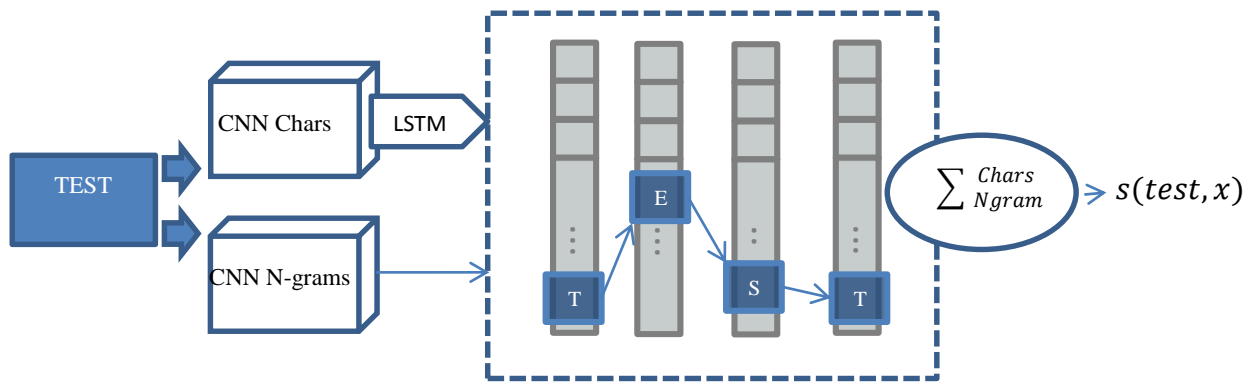

Figure 7:1 The joint-LSTM model combines CNN Chars + LSTM and n-gram models

We also plan to make use of the whole synthetic dataset with more powerful computers. Besides we would like to train the CNN and the LSTM models simultaneously instead of training the CNN and the LSTM independently. Moreover, we would like to use a lexicon after the LSTM predictions, as shown in figure 7.2, and compare it against the lexicon based approaches' performance.

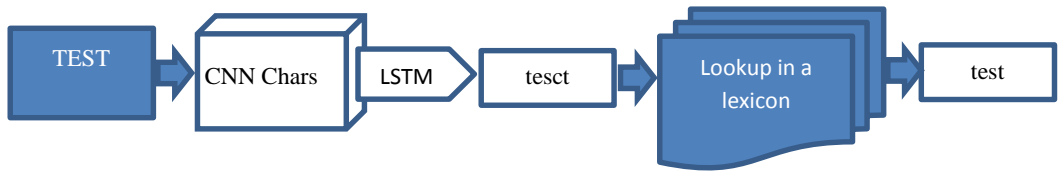

Figure 7:2 The joint-LSTM model combines CNN Chars + LSTM and n-gram models

# Appendix A

We present in this section some of the experimented models as defined in our code.

Model no. 1.
```python
peepholes = True
        l_in = lasagne.layers.InputLayer(shape=(BATCH_SIZE, MAX_SEQ_LENGTH,
N_FEATURES))
        recout = lasagne.layers.BidirectionalLSTMLayer(
            l_in, num_units=102, peepholes=peepholes, learn_init=True)
        recout = lasagne.layers.BidirectionalLSTMLayer(
            recout, num_units=156, peepholes=peepholes, learn_init=True)
        l_reshape = lasagne.layers.ReshapeLayer(
            recout,  (BATCH_SIZE*MAX_SEQ_LENGTH, recout.get_output_shape()[-1]))
        l_rec_out = lasagne.layers.DenseLayer(
```

```
            l_reshape, num_units=N_CLASSES,
nonlinearity=lasagne.nonlinearities.softmax)
        l_out = lasagne.layers.ReshapeLayer(
    l_rec_out, (BATCH_SIZE, MAX_SEQ_LENGTH, N_CLASSES))
```

The best performing LSTM model architecture. Model no 17

```
peepholes = True
l_in = lasagne.layers.InputLayer(shape=(BATCH_SIZE, MAX_SEQ_LENGTH,
    N_FEATURES))
recout_1 = lasagne.layers.BidirectionalLSTMLayer(l_in, num_units=450, peep
    holes=peepholes, learn_init=True)
recout_2 = lasagne.layers.BidirectionalLSTMLayer(recout_1, num_units=400,
    peepholes=peepholes, learn_init=True)
recout_2_dropout = lasagne.layers.DropoutLayer(recout_2, p=0.5)
recout_3 = lasagne.layers.BidirectionalLSTMLayer(recout_2_dropout,
    num_units=350, peepholes=peepholes, learn_init=True)
recout_3_dropout = lasagne.layers.DropoutLayer(recout_3, p=0.5)
recout_4 = lasagne.layers.BidirectionalLSTMLayer(recout_3_dropout,
    num_units=300, peepholes=peepholes, learn_init=True)
recout_4_dropout = lasagne.layers.DropoutLayer(recout_4, p=0.5)
l_reshape = lasagne.layers.ReshapeLayer(recout_4_dropout,
    (BATCH_SIZE*MAX_SEQ_LENGTH, recout_4_dropout.get_output_shape()[-1]))
l_rec_out = lasagne.layers.DenseLayer(l_reshape, num_units=N_CLASSES, nonline
    arity=lasagne.nonlinearities.softmax)
l_out = lasagne.layers.ReshapeLayer(l_rec_out, (BATCH_SIZE, MAX_SEQ_LENGTH,
N_CLASSES))
```

# Appendix B

# Datasets

In this work we mainly usde two datasets, the ICDAR 2013 [33] and the synthetic dataset [12]. We evaluate our approach on ICDAR 2013 word recognition test set which consists of 1439 images. The task is focused on the reading of text in real scenes.

The figure 7:2 shows a random sample of the test set images in their original size.

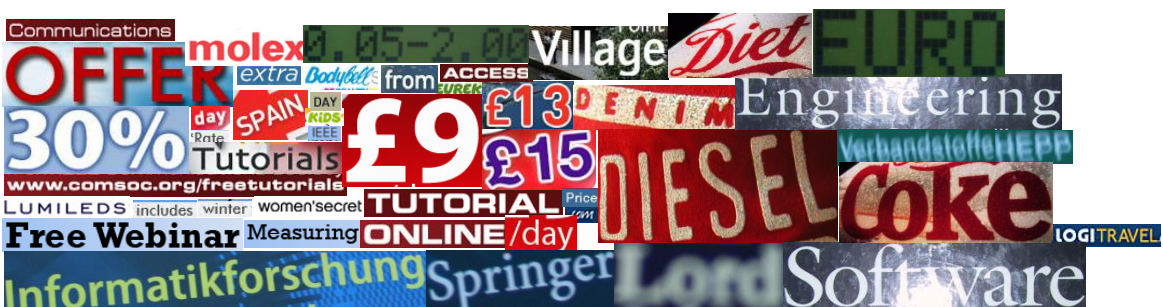

Figure 7:3 ICDAR 2013 test set sample

The synesthetic dataset is a part of the visual geometry group work [18]. The dataset consists of 9M images covering 90K English words. The figure below shows a sample of the synthetic dataset

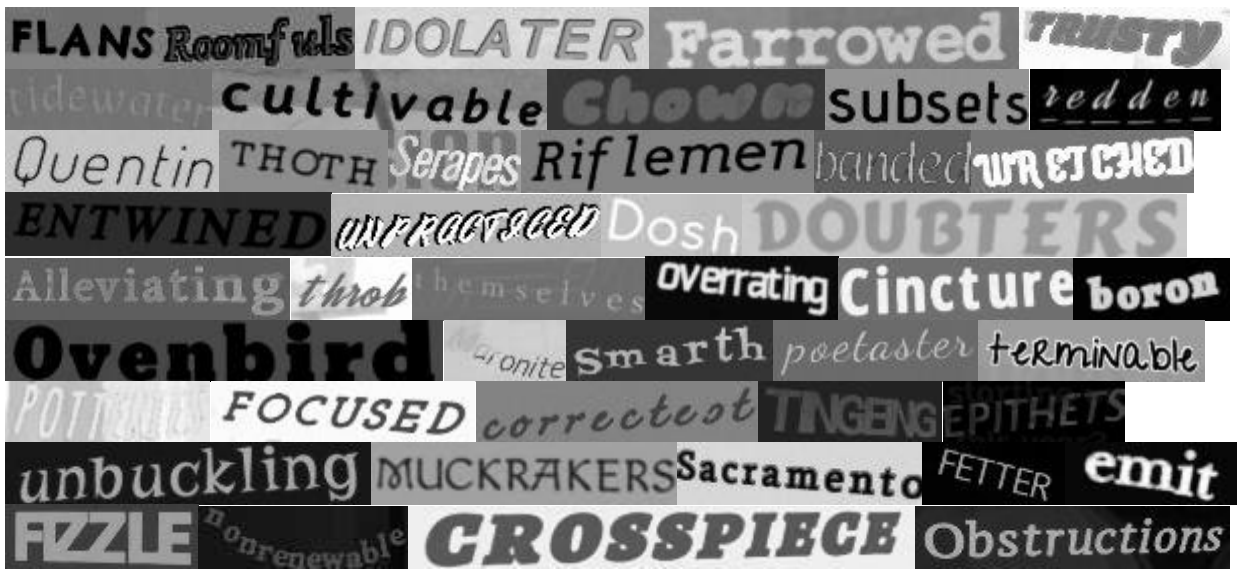

Figure 7:4 Random sample of the synthetic dataset.